\documentclass[11pt]{article} 
\usepackage{amsmath}
\usepackage{lscape}
\usepackage[section]{placeins}
\usepackage{mathtools}
\usepackage{amsthm,bm}
\usepackage{times}
\usepackage{graphicx}
\usepackage{amssymb}
\usepackage{url} 
\usepackage{microtype}
\usepackage{authblk}
\usepackage{siunitx}
\usepackage{subcaption}
\usepackage[margin=1in]{geometry}
\usepackage{longtable}
\usepackage{booktabs}
\usepackage{multirow}
\usepackage{xcolor}
\usepackage{comment}
\usepackage{color,soul} 
\usepackage{graphicx} 
\usepackage{caption}
\usepackage{algorithm}
\usepackage{algpseudocode}
\usepackage{booktabs} 
\usepackage{float}
\usepackage{minitoc}

\usepackage[toc,page,header]{appendix}
\usepackage{etoc}
\definecolor{darkgreen}{RGB}{0,90,0.1}
\usepackage[colorlinks=true, citecolor=darkgreen, urlcolor=darkgreen]{hyperref}
\usepackage{multicol}
\usepackage{dcolumn}
\usepackage{enumitem}
\usepackage[capitalize,noabbrev]{cleveref}
\usepackage[version=4]{mhchem} 
\usepackage[textsize=tiny]{todonotes}
\usepackage{natbib}
\usepackage{tikz}
\usetikzlibrary{calc}
\usetikzlibrary{arrows.meta,positioning,fit,backgrounds,shapes.geometric,  decorations.pathreplacing}

\hypersetup{
    colorlinks=true,
    linkcolor=blue,
    urlcolor=blue,
    citecolor=red
}

\theoremstyle{plain}
\newcommand{\Lcal}{\mathcal{L}}
\newtheorem{theorem}{Theorem}[section]
\newtheorem{proposition}[theorem]{Proposition}
\newtheorem{lemma}[theorem]{Lemma}

\theoremstyle{definition}

\theoremstyle{remark}
\newtheorem{remark}[theorem]{Remark}

\title{RRaPINNs:  Residual Risk-Aware  Physics Informed Neural Networks}
\begin{document}
\author[1,2]{ Ange-Clément Akazan}
\author[2]{ Issa Karambal }
\author[1]{ Jean Medard Ngnotchouye}
\author[3]{ Abebe Geletu Selassie. W}

\affil[1]{University of KwaZulu Natal}
\affil[2]{AIMS Research and Innovation Centre}
\affil[3]{AIMS Rwanda}
\maketitle

\begin{abstract}
Physics-informed neural networks (PINNs) typically minimize average residuals, which can conceal large, localized errors. We propose \emph{Residual Risk-Aware PINNs} (RRaPINNs), a single-network framework that optimizes tail-focused objectives using Conditional Value-at-Risk (CVaR), we also introduced a  Mean-Excess (ME) surrogate penalty  to directly control worst-case PDE residuals. This casts PINN training as risk-sensitive optimization and links it to chance-constrained formulations. The method is effective  and simple to implement. Across several partial differential equations (PDEs)  such as Burgers, Heat, Korteweg-de-Vries, and Poisson (including a Poisson interface problem with a source jump at x=0.5) equations, RRaPINNs reduce tail residuals while maintaining or improving mean errors compared  to vanilla PINNs, Residual-Based Attention and its variant using convolution weighting; the ME surrogate yields smoother optimization than a direct CVaR hinge. The chance constraint reliability level $\alpha$ acts as a transparent knob trading bulk accuracy (lower $\alpha$ ) for stricter tail control (higher $\alpha$ ). We discuss the framework limitations, including memoryless sampling, global-only tail budgeting, and residual-centric risk, and outline remedies via persistent hard-point replay, local risk budgets, and multi-objective risk over BC/IC terms. RRaPINNs offer a practical path to reliability-aware scientific ML for both smooth and discontinuous PDEs.
\end{abstract}

\section{Introduction}

Physics-Informed Neural Networks (PINNs), a framework introduced by \citep{Lagaris_1998} and later on improved by \citep{raissi2019pinns}, have emerged as a powerful paradigm for solving partial differential equations (PDEs) by embedding physical laws into neural network training. By replacing or augmenting data-driven losses with PDE residuals and boundary/initial conditions, PINNs provide a mesh-free framework for scientific machine learning, with demonstrated success across fluid dynamics, elasticity, and transport phenomena \cite{karniadakis2021physics}.

However, as first demonstrated by \citet{rahaman2019spectral},  neural networks inherently learn low-frequency functions (the smooth, easy parts of a solution) far more quickly than high-frequency functions (sharp gradients, shocks, or singularities). The classical PINN formulation, which minimizes the \emph{mean-squared residual}, exacerbates this problem. The optimizer rapidly reduces the low-frequency error across the bulk of the domain, causing the global average loss to drop. This  averaging  obscures the localized failures in high-frequency regions, leading to a false sense of convergence while the solution at these critical points remains catastrophically wrong \citep{wang2022understanding,article_survey}. 
Moreover, ill-conditioned physics-informed losses, can  also yield highly unbalanced gradient signals across the domain \citep{rahaman2019spectral, wang2021understanding}.
Together, these often  lead  \emph{non-Gaussian, heavy-tailed residual distribution}, where the standard MSE loss is no longer a statistically robust or meaningful objective. Such failure modes undermine the reliability of PINNs, especially in safety-critical domains.

Adaptive weighting schemes have been proposed to mitigate this pathology. These include balancing PDE vs. boundary losses \citep{mcclenny2020adaptive}, gradient norm balancing \citep{wang2021understanding}, adaptive sampling of collocation points \citep{lu2021deepxde}, and curriculum-based strategies \citep{yang2022cwp}. These methods mainly attempt to force the optimizer to pay attention to these high-frequency, high-residual regions. Despite their successes, these approaches remain largely heuristic. What is missing is a systematic, mathematically rigorous framework that directly targets the statistical properties of this heavy-tailed error distribution. Addressing this requires moving beyond simple reweighting toward \emph{robust error control}. In this work, we propose a framework which formulates the training problem as a \emph{probabilistic (chance)-constrained optimization} \citep{charnes1959chance}.
Rather than minimizing average residuals only, we also enforce a probabilistic constraint on the residual.
\begin{equation}
    \mathbb{P}\big(|r(x;\theta)| \leq \varepsilon\big) \geq \alpha,
    \label{eq:chance_constraint}
\end{equation}
where $r(x;\theta)$ is the PDE residual, $\varepsilon$ a tolerance, and $\alpha$ the desired confidence level (e.g.\ 95\%). 

Beyond heuristic reweighting, our view is to re-frame PINN training as a probabilistic (chance)-constrained problem: we require that a large fraction ($\alpha$) of collocation points meet a at most a residual tolerance $\epsilon$. This formulation naturally induces a quantile-driven objective, targeting the $(1-\alpha)$-quantile (also known as Value-at-Risk or VaR) of the residual distribution. However, quantiles are ill-suited for gradient-based optimization. As a risk measure, VaR is non-coherent; as an objective, it is piecewise-constant, yielding zero gradients almost everywhere and providing sparse, unstable signals to modern optimizers.We therefore turn to \emph{Conditional Value-at-Risk} (CVaR), the expected loss given that the loss exceeds the VaR. CVaR is a coherent, convex, and smoothly optimizable measure of tail risk \citep{rockafellar2000optimization}. Crucially, it provides dense, informative gradients from the entire tail, not just a single point. This perspective is central to modern risk-averse and distributionally robust learning, which use tail-aware objectives to harden models against rare but consequential errors \citep{namkoong2017variance,duchi2018learning}. We bring this risk-averse perspective directly to the physics loss in PINNs, thereby introducing \textbf{\emph{Residual Risk-Aware PINNs (RRaPINNs)}}. In addition to the physics loss, this framework penalizes the positive excess of the residual tail beyond an \emph{adaptive} tolerance $\epsilon$, allowing the optimizer to explicitly target and shape the worst $(1-\alpha)$ slice of the residual distribution. 

To the best of our knowledge, our risk-based formulation is the first to elevate PINN training from heuristic reweighting to a principled, distributionally robust framework that unifies precision with reliability. 
We motivate and apply coherent risk measures (CVaR) directly to the PDE residual loss to explicitly model and control the upper $(1-\alpha)$ tail of the error distribution. We provide a theoretical derivation showing that this CVaR formulation induces a principled, data-dependent reweighting of residuals.  We architect a novel and  stable composite-loss framework (RRaPINN). Instead of a naive two-stage switch, we augment the standard MSE base loss with our CVaR-based penalty. To solve the critical problem of loss-scale mismatch, a primary source of instability in high-order PDEs, we introduce a dynamic loss-balancing mechanism. This key  adaptively tunes the penalty weight ($\lambda_p$) using the ratio of detached, exponentially-moving-average (EMA) scales of the base and penalty losses. This ensures a stable, multi-objective optimization of both the average and worst-case residuals without manual tuning. We demonstrate the efficacy of RRaPINN on a comprehensive benchmark of five PDE problems. Our method consistently  outperforms Vanilla (baseline) PINNs \citep{raissi2019pinns}, residual based attention \citet{ANAGNOSTOPOULOS2024116805} and convolution weighting PINN \citep{si2025convolutionweightingmethodphysicsinformedneural} in controlling not only the average relative $L^2$ error but also the $L^\infty$ norm and high-percentile residual errors. We provide definitive visual evidence via an empirical Complementary Cumulative Distribution Function (CCDF) plots, which explicitly show our method's ability to pull in and suppress the extreme tail of the residual distribution.

\medskip

The remainder of this work is structured as follows: section \ref{background} describes the background of this research, section \ref{r_w} discusses some of the most important related works. The next section \ref{sec:method} discusses the methodology adopted in this research. Section \ref{experiments} describes and discusses the  experimental results and investigates  the impact of the tail size parameter $\alpha$ on accuracy and residual tail control. 

\section{Background}
\label{background}
This section provides the background of this research, describing the PINNs initial framework.
\subsection{PINNs and the Challenge of Adaptive Weighting}

\paragraph{Problem Setup.}
Let $\Omega\subset\mathbb{R}^d$ be a spatial domain and $[0, T]$ a time interval. The full spatio-temporal domain is $\mathcal{D} = \Omega \times (0, T]$. The boundary is composed of the spatial boundary $\partial\Omega \times [0, T]$ and the initial-time boundary $\Omega \times \{0\}$. We consider a parameterized PDE:
\begin{equation}
\mathcal{N}[u](x, t) = f(x, t), \quad (x, t) \in \mathcal{D}, 
\label{eq:pde-main}
\end{equation}
subject to boundary conditions $\mathcal{B}$ and initial conditions $\mathcal{I}$:
\begin{equation}
\mathcal{B}[u](x, t) = g(x, t), \quad (x, t) \in \partial\Omega \times [0, T],
\qquad 
\mathcal{I}[u](x, 0) = h(x), \quad x \in \Omega.
\label{eq:pde-bvp}
\end{equation}
A neural network $u_\theta(x, t)$ with parameters $\theta$ is used to approximate the solution $u$. The training objective is defined by minimizing three terms: the \emph{PDE residual}, the \emph{boundary discrepancy}, and the \emph{initial discrepancy}:
\begin{align}
r(x, t;\theta) &:= \mathcal{N}[u_\theta](x, t) - f(x, t), \label{eq:res-r}\\
b(x, t;\theta) &:= \mathcal{B}[u_\theta](x, t) - g(x, t), \label{eq:res-b}\\
i(x;\theta) &:= \mathcal{I}[u_\theta](x, 0) - h(x). \label{eq:res-i}
\end{align}
Let $\mu_r, \mu_b, \mu_i$ be sampling measures on the corresponding domains. The classical (vanilla) PINN objective minimizes the empirical $L_2$ loss over the defined  sampling measures:
\begin{equation}
\widehat{\Lcal}_{\text{PINN}}(\theta) 
= \lambda_r  \Lcal_r(\theta) + \lambda_b \Lcal_b(\theta) + \lambda_i \Lcal_i(\theta),
\label{eq_vanilla_pinn} 
\end{equation}
where
\begin{align}
\Lcal_r(\theta) &= \frac{1}{N_r}\sum_{j=1}^{N_r} r(x_j^r, t_j^r;\theta)^2,  
\Lcal_b(\theta) = \frac{1}{N_b}\sum_{j=1}^{N_b} b(x_j^b, t_j^b;\theta)^2,   \text{and}   
\Lcal_i(\theta) = \frac{1}{N_i}\sum_{j=1}^{N_i} i(x_j^i;\theta)^2.
\end{align}

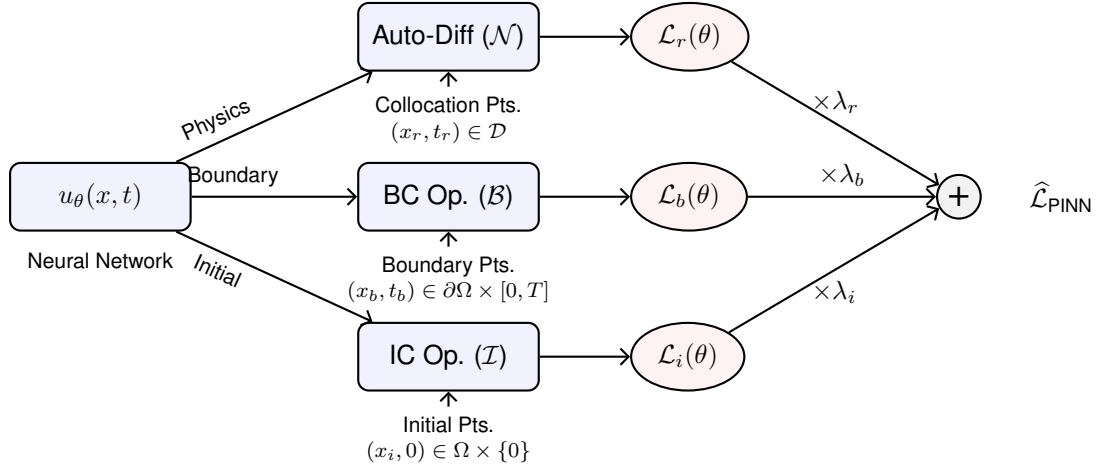
\begin{figure}[htbp]
\centering
\begin{tikzpicture}[
    font=\sffamily\small,
    node distance=2cm and 1.2cm,
    box/.style={draw, rectangle, rounded corners, thick, minimum height=0.9cm, minimum width=2.4cm, fill=blue!5},
    loss/.style={draw, ellipse, thick, minimum height=0.9cm, fill=red!5},
    sum/.style={draw, circle, thick, font=\Large, inner sep=2pt, fill=gray!10},
    arrow/.style={-Straight Barb, thick},
    label/.style={align=center, font=\sffamily\scriptsize}
]
    \node[box] (nn) {$u_\theta(x, t)$};
    \node[label, below=0.15cm of nn] (nn_label) {Neural Network};
    
    \node[box, above right=1.2cm and 2.2cm of nn] (ad_pde) {Auto-Diff ($\mathcal{N}$)};
    \node[loss, right=of ad_pde] (loss_r) {$\Lcal_r(\theta)$};
    \node[label, below=0.2cm of ad_pde] (input_r) {Collocation Pts. \\ $(x_r, t_r) \in \mathcal{D}$};
    \draw[arrow] (input_r.north) -- (ad_pde.south);
    
    \node[box, right=2.2cm of nn] (ad_bc) {BC Op. ($\mathcal{B}$)};
    \node[loss, right=of ad_bc] (loss_b) {$\Lcal_b(\theta)$};
    \node[label, below=0.2cm of ad_bc] (input_b) {Boundary Pts. \\ $(x_b, t_b) \in \partial\Omega \times [0,T]$};
    \draw[arrow] (input_b.north) -- (ad_bc.south);

    \node[box, below right=1.2cm and 2.2cm of nn] (ad_ic) {IC Op. ($\mathcal{I}$)};
    \node[loss, right=of ad_ic] (loss_i) {$\Lcal_i(\theta)$};
    \node[label, below=0.2cm of ad_ic] (input_i) {Initial Pts. \\ $(x_i, 0) \in \Omega \times \{0\}$};
    \draw[arrow] (input_i.north) -- (ad_ic.south);
    
    \node[sum, right=2.5cm of loss_b] (total_loss) {+};
    \node[right=0.3cm of total_loss, xshift=2mm] (loss_label) {$\widehat{\Lcal}_{\text{PINN}}$};

    \draw[arrow] (nn) -- node[above, near start, sloped, font=\sffamily\scriptsize] {Physics} (ad_pde);
    \draw[arrow] (nn) -- node[above, near start, font=\sffamily\scriptsize] {Boundary} (ad_bc);
    \draw[arrow] (nn) -- node[below, near start, sloped, font=\sffamily\scriptsize] {Initial} (ad_ic);
    
    \draw[arrow] (ad_pde) -- (loss_r);
    \draw[arrow] (ad_bc) -- (loss_b);
    \draw[arrow] (ad_ic) -- (loss_i);
    
    \draw[arrow] (loss_r) -- node[above, font=\sffamily\small] {$\times \lambda_r$} (total_loss);
    \draw[arrow] (loss_b) -- node[above, font=\sffamily\small] {$\times \lambda_b$} (total_loss);
    \draw[arrow] (loss_i) -- node[below, font=\sffamily\small] {$\times \lambda_i$} (total_loss);
    
\end{tikzpicture}
\caption{Computational graph of a standard PINN. The neural network $u_\theta$ is evaluated at different sets of collocation points and processed through corresponding operators (PDE, BC, IC) to compute distinct loss terms. These terms are weighted (multiplied) and summed to form the final training objective $\widehat{\Lcal}_{\text{PINN}}$.}
\label{fig:pinn_graph}
\end{figure}
The static weights $(\lambda_r, \lambda_b, \lambda_i)\in \mathbb{R}_+^3$, are notoriously difficult to tune. A naive choice (e.g., all equal to 1) often leads to failed training due to the pathological nature of the PINN loss landscape. The core issue lies in the disparate gradient magnitudes from each loss term. The residual loss $\Lcal_r$, which involves high-order derivatives, often produces gradients that are orders of magnitude stiffer (or vanishingly small) than the simple data-fitting losses $\Lcal_b$ and $\Lcal_i$. Furthermore, the spectral bias of neural networks~\citep{rahaman2019spectral} causes the model to rapidly learn low-frequency boundary conditions while struggling to fit the high-frequency components of the PDE residual. This imbalance creates a conflict where one loss term dominates the optimization, leading to a converged model that may satisfy the boundary conditions but violates the governing physics, or vice versa~\citep{wang2021understanding,wang2022ntk}.
To alleviate this issue, several adaptive strategies have been proposed, including
\emph{adaptive weighting}, \emph{adaptive sampling}, and \emph{curriculum learning} approaches that dynamically redistribute the training effort across loss terms or regions of
the domain.

\section{Related Works }
\label{r_w}
This section discusses the  main adaptive strategies  that dynamically redistribute the training effort across loss terms or regions of
the domain.
\paragraph{Adaptive Weighting Strategies in PINNs}
Balancing heterogeneous loss components remains a core challenge in Physics-Informed Neural Networks (PINNs), where PDE residuals, boundary, and initial terms may differ by orders of magnitude \cite{Lagaris_1998,raissi2019pinns}. Without correction, training often suffers from gradient domination or vanishing updates in certain constraints. 
The adaptive weighting methods have emerged to address this imbalance.
\emph{Gradient-based schemes.}
\citet{wang2021understanding} proposed a GradNorm-inspired attention mechanism that rescales losses based on gradient statistics, enabling the optimizer to emphasize under-trained terms. While effective in early training, such schemes introduce noise sensitivity and require additional tuning to prevent oscillatory reweighting. 
\emph{NTK-based weighting}
\cite{wang2022ntk} derive the NTK-PINN  and show that mismatched NTK spectra across loss components (e.g., PDE residual vs.\ boundary terms) cause unequal convergence rates. They propose an adaptive scheme that \emph{updates the loss coefficients} $\{\lambda_i\}$ using spectral information from the corresponding NTK blocks (e.g., traces/eigenvalue magnitudes), so that each component’s error decays at a comparable rate. The approach improves balance but requires computing (or approximating) NTK spectral quantities during training, which limits scalability.
\emph{Self-Adaptive Schemes.}
\cite{mcclenny2023selfadaptive} proposed SA-PINNs, where trainable pointwise weights act as soft-attention masks to focus learning on high-residual regions. This joint ascent-descent optimization improves convergence on stiff PDEs and balances loss terms through localized adaptivity. However, the method is sensitive to learning-rate tuning and lacks a principled probabilistic basis, making it largely heuristic despite its clear practical gains.
\emph{Residual-based attention.}
To bypass gradient computations, \citet{ANAGNOSTOPOULOS2024116805} introduced a residual-based attention (RBA) mechanism that adjusts weights as bounded functions of cumulative residuals. RBA accelerates convergence and highlights difficult regions, though its heuristic design lacks explicit guarantees on tail-residual control. \emph{Convolution-weighting Scheme.}
More recently, \cite{si2025convolutionweightingmethodphysicsinformedneural} introduced CWP a  method that smooths residuals through local convolutions, enforcing spatial coherence in adaptive weights. This primal-dual formulation mitigates overfitting to isolated points and improves convergence on stiff PDEs. While effective and theoretically grounded, it depends on kernel hyperparameters and may oversmooth sharp features, limiting flexibility in highly nonlocal or discontinuous problems.

\paragraph{Adaptive Collocation Points Sampling in PINNs.}
Adaptive resampling aims to reallocate collocation points toward regions where the PDE residual remains large, thus improving solution fidelity in underresolved or shock regions. \cite{lu2021deepxde} introduced DeepXDE which fosters residual-based refinement, by iteratively adding points with the highest residuals to enhance local accuracy. 
\emph{Failure-Informed Adaptive Sampling (FI-PINNs).}
\citet{gao2023failureinformedadaptivesamplingpinns} proposed the
\emph{Failure-Informed PINN} (FI-PINN), which defines a failure probability as the likelihood that the PDE residual exceeds a user-defined threshold. When this probability surpasses a tolerance, new collocation points are added in the high-residual (failure) region using a
self-adaptive importance sampling scheme based on truncated Gaussian proposals. The method provides rare theoretical error bounds for adaptive PINNs and achieves superior accuracy on stiff and high-dimensional PDEs. Its main limitations lie in the need to tune the residual and probability thresholds combined with its reliance on Gaussian assumptions, which may underperform in highly anisotropic or multimodal residual landscapes.
\citet{wu2023residualsampling} extended the  residual-based refinement by formulating sampling as a probability distribution, where each point is drawn with likelihood proportional to its normalized residual magnitude. This probabilistic view avoids redundant sampling and improves efficiency when the sampling budget is limited. More recently, \citet{si2025convolutionweightingmethodphysicsinformedneural} coupled resampling with convolution-weighted training, using smoothed residuals to guide point updates, thereby ensuring spatial consistency. While these methods significantly improve convergence and robustness, most rely on heuristic thresholds or empirical schedules and lack formal convergence guarantees or uncertainty-aware sampling criteria.

\paragraph{Curriculum Methods.}
Curriculum strategies aim to ease PINN optimization by introducing physical constraints or domain complexity gradually rather than enforcing all conditions simultaneously. In this spirit,  \citet{yang2022cwp} proposed the \emph{Curriculum PINN} (C-PINN), where training begins with simpler tasks, such as satisfying boundary conditions or low-frequency
components, and progressively includes full PDE residuals as the model stabilizes. \citet{krishnapriyan2021characterizing} highlighted that premature enforcement of difficult regions often leads to convergence failures, motivating curricula that schedule constraint difficulty. \cite{jagtap2021xpinns} addressed
similar issues via domain decomposition (\emph{XPINNs}), effectively providing a spatial curriculum that trains subdomains before coupling them. More recently, \citet{wang2022causalpinn} proposed \emph{Causal PINNs}, where temporal weights enforce causality by prioritizing earlier time segments, improving stability for transient problems. Overall, curriculum approaches stabilize training and accelerate convergence but depend on heuristic scheduling and problem-specific tuning, lacking general criteria for optimal
curriculum design.

\medskip

\paragraph{Risk-averse neural operators}
Related work applies CVaR to \emph{neural operators} to improve out-of-distribution generalization on supervised, function-to-function tasks \citep{10.1145/3711896.3737020}. Our setting differs along two axes.
(i) PINNs solve a specific PDE by enforcing the governing equations and boundary/initial conditions, rather than learning a global operator from paired functions.
(ii) we apply risk directly to the \emph{physics residual} and can train without labeled solutions (physics-only), whereas neural operators apply risk to \emph{supervised prediction losses} and require function-pair datasets.
As a result, the algorithms, training dynamics, and evaluation criteria diverge; our approach complements NO-based OOD robustness.

\medskip

These PINN-based approaches  share a common premise: dynamically adjusting the effective weights assigned to loss terms or training points to balance learning across the domain. 

Despite their empirical success,  these methods do not fundamentally alter the $L_2$ (MSE) objective itself, which remains risk-neutral and highly sensitive to outliers.
They also remain mostly \emph{heuristic}, the weighting rules are
often ad hoc (e.g., proportional to gradient magnitudes or residual norms) and lack a  rigorous optimization-theoretic coupled with probabilistic foundation. Moreover, none of these strategies incorporate an explicit \emph{uncertainty-aware mechanism} to characterize or control the distribution of residual errors. Consequently, they provide no guarantees that the trained model achieves uniformly reliable accuracy across the domain. In fact, they often demonstrate systematic underperformance in rare but critical regions, such as shocks, singularities, or stiff regimes, where large residuals persist despite apparent global
convergence.


\section{Residual Risk-Aware PINN (RRaPINN) Methodology}
\label{sec:method}
This part provides a structured methodology starting by elaborating our PINNs-basedprobabilistic framework, the benefit of using the CVaR instead of the VaR probabilistic constraint. We then jump into the main RRaPINN formulation based on theoretical analogies between CVaR and the top-k loss. We also provided some training tips to improve our methods, and the metrics used in this study.
\subsection{ From Mean Errors to Probabilistic Guarantees}
The objective \eqref{eq_vanilla_pinn} controls only the \emph{mean} residual and can hide large localized errors, which may deteriorate the approximation performance. To resolve this, we introduce a distributional guarantee on the absolute residual which is considered as a  random variable given random collocation points.
\begin{equation}
R_{\theta} := |r(X;\theta)|, \qquad X\sim\mu,
\label{eq:R-rv}
\end{equation}
via the \emph{chance constraint}
\begin{equation}
\mathbb{P}\Big(R_{\theta} \le \varepsilon\Big)  \ge  \alpha,
\qquad \text{for some } \alpha\in (0.5,1),\ \varepsilon>0.
\label{eq:chance_constraint_1}
\end{equation} 
Thus, \eqref{eq_vanilla_pinn} becomes which gives the following probabilistic constrained problem:
\begin{align}
\min_{\theta} \quad & 
\mathcal{L}_{\text{PINN}}(\theta)
\label{eq:base-obj_}\\[2pt]
\text{s.t.} \quad &
\mathbb{P}\Big(R_{\theta} \le \varepsilon\Big)  \ge  \alpha
\label{pco_pinn}
\end{align}
$\varepsilon$ can be either learned or adaptive  while allowing the model to choose a tolerance consistent with the physics and capacity).

\subsubsection{Quantile Formulation (Value-at-Risk).}
Considering the residual magnitudes $R_\theta:=|r(X;\theta)|$,  let $F_\theta(t):=\mathbb{P}(R_\theta\le t)$ denote the CDF of $R_\theta$ , and let 
$Q_{R_{\theta}}(\alpha):=\inf\{t\in\mathbb{R}^+:F_{\theta}(t)\ge \alpha\}$ be its $\alpha$-quantile. As demonstrated in \citep{nemirovski2007convex},
the chance constraint \eqref{eq:chance_constraint_1} is therefore equivalent to
\begin{equation}
\mathrm{VaR}_{\alpha}(R_\theta) := Q_{R_\theta}(\alpha) \leq \varepsilon,
\label{eq:var-def}
\end{equation}
where $\mathrm{VaR}_{\alpha}$ (Value-at-Risk) \citep{duffie1997overview} is the classical surrogate risk measure widely used in quantitative finance. 
For $\alpha\in(0,1)$, $\mathrm{VaR}_\alpha(R_\theta)$ specifies the threshold below which at least a fraction $\alpha$ of \textit{residuals} lie. 
While this captures where the ``worst" $(1-\alpha)$ proportion of the distribution begins, it says nothing about the magnitude of those extreme values. 
Moreover, empirical estimates of VaR are \emph{piecewise-constant} and hence non-differentiable with respect to $\theta$, so their gradients vanish almost everywhere and are unstable near order-statistic ties. 
VaR also fails to satisfy key axioms of a coherent risk measure, such as sub-additivity. 
These limitations make it ill-suited as a training objective and motivate the use of a smoother, tail-sensitive alternative: the \emph{Conditional Value-at-Risk (CVaR)} \citep{rockafellar2000optimization}.

\subsubsection{Conditional Value-at-Risk (CVaR).}
The Conditional Value-at-Risk refines VaR by measuring the \emph{expected loss in the tail}, i.e., the average of outcomes that exceed the VaR threshold:
\begin{equation}
\mathrm{CVaR}_\alpha(R_\theta)=\mathbb{E}\big[R_\theta \big| R_\theta \ge \mathrm{VaR}_\alpha(R_\theta)\big].
\end{equation}
Unlike VaR, CVaR is a \emph{coherent} risk measure,  convex,  and explicitly sensitive to the severity of extreme losses. 
This property makes CVaR particularly suitable in the PINN setting, where collocation regions often contain localized uncertainties, shocks, or singularities that can lead to disproportionately large residuals. 
By directly penalizing the average severity of these extreme events, CVaR provides a principled, optimization-friendly surrogate for robust residual control. Therefore our robust formulation can be written as :
\begin{equation}
\min_{\theta,\varepsilon>0}   \underbrace{\mathcal{L}_{\text{PINN}}(\theta)}_{\text{physics + boundary}}
\quad  \text{s.t.} \quad 
\underbrace{\mathrm{CVaR}_\alpha\big(|r(X;\theta)|\big)}_{\text{tail risk of residuals}}
\le \varepsilon,
\label{pco_pinn_cvar}
\end{equation}

As \citet{rockafellar2000optimization} demonstrated, the \emph{Conditional Value-at-Risk} at the confidence level $\alpha$ can be rewritten as the Rockafellar-Uryasev (RU) program as follows\begin{align}
\mathrm{CVaR}_{\alpha}(R)
&:= \inf_{\eta\in\mathbb{R}}   \phi(\eta),
\qquad 
\phi(\eta)= \eta + \frac{1}{1-\alpha}\mathbb{E}\big[(R-\eta)_+\big],
\label{eq:cvar-ru}\\
(x)_+ &:= \max\{x,0\}.
\end{align}


The empirical CVAR  RU objective is defined as follows:
\begin{equation}
\widehat{\phi}_N(\eta) = \eta + \frac{1}{1-\alpha}\cdot \frac{1}{N}\sum_{i=1}^N (R_i-\eta)_+.
\label{eq:emp-ru}
\end{equation}
Unlike VaR, $\phi(\eta)$ is convex and (sub)differentiable in both $\eta$ and in the neural network's parameters $\theta$, enabling stable stochastic optimization.

\subsection{  From Empirical CVAR  to Residual Adaptive Weighting View.}
\subsubsection{Top-k Set Definition}
\label{subsec:topk-structure}

Given a randomly sampled  minibatch $\{x_i\}_{i=1}^N$, define residual magnitudes
\(
R_i(\theta) := |r(x_i;\theta)|, \quad i=1,\dots,N.
\) Let the order statistics be $R_{(1)}(\theta)\le\cdots\le R_{(N)}(\theta)$ and let $\pi$ be a permutation such that $R_{(\ell)}(\theta)=R_{\pi(\ell)}(\theta)$. Fix $k:=\lfloor N(1-\alpha)\rfloor$ and define the \emph{Top-$k$ index set}
\begin{equation}
    \mathcal{I}_{\text{top-}k}(\theta) := \big\{\pi(N-k+1),\dots,\pi(N)\big\}.
\end{equation}
The associated \emph{threshold} (the smallest value included in the Top-$k$) is \(
\tau(\theta) := R_{(N-k+1)}(\theta).
\)
\paragraph{Ties at the threshold.}
Let $\mathcal{E}(\theta):=\{i: R_i(\theta)=\tau(\theta)\}$ denote indices equal to the threshold. If there are ties at $\tau(\theta)$, the set $\mathcal{I}_{\text{top-}k}(\theta)$ is not unique; any selection with cardinality $k$ that includes all strictly larger values and an arbitrary subset of $\mathcal{E}(\theta)$ of the required size is valid. All such selections yield the same empirical CVaR value.
\begin{remark}
    All the ascendant residual order statistics ($R_{(1)}(\theta)\le\cdots\le R_{(N)}(\theta)$) provided in this study are training-specific and may vary from one epoch to the other.
\end{remark}
\begin{proposition}[Proof\footnote{The proof of the proposition ~\eqref{prop:emp-cvar-detailed}  can be found in the Appendix ~\ref{lem:phi-subdiff} }]
\label{prop:emp-cvar-detailed} 
Let $R_{(1)}(\theta)\le\cdots\le R_{(N)}(\theta)$ be order statistics and set $t=(1-\alpha)N$, $m=\lfloor t\rfloor$ ($\lfloor t\rfloor$ being the floor of $t$), $s=t-m$.
Then the empirical RU objective
\begin{equation}
\widehat{\phi}_N(\eta)=\eta+\frac{1}{t}\sum_{i=1}^N (R_i-\eta)_+    
\end{equation} has minimizers
\begin{equation}
    \arg\min_\eta \widehat{\phi}_N(\eta)\in
\begin{cases}
\{R_{(N-m)}\}, & \text{if $s>0$ and there are no ties at $R_{(N-m)}$},\\
[R_{(N-m)},R_{(N-m+1)}], & \text{if $s=0$ or there is a tie at the boundary}.
\end{cases}
\end{equation}
\end{proposition}
\begin{proposition}[Proof  \footnote{The proof of the proposition  ~\eqref{prop_emp-cvar-fractional} can be found in the Appendix ~\ref{lem:phi-subdiff} }] 
\label{prop_emp-cvar-fractional} 
Let $R_1,\dots,R_N\in\mathbb{R}$ and let $R_{(1)}\le \cdots \le R_{(N)}$ be the order statistics (ascending).
Fix $\alpha\in(0,1)$ and set $t:=(1-\alpha)N$, $m:=\lfloor t\rfloor$, $s:=t-m\in[0,1)$.
Consider the empirical Rockafellar-Uryasev objective
\[
\widehat{\phi}_N(\eta)=\eta + \frac{1}{(1-\alpha)N}\sum_{i=1}^N (R_i-\eta)_+
= \eta + \frac{1}{t}\sum_{i=1}^N (R_i-\eta)_+.
\]
Then the minimum value (the empirical CVaR) is
\begin{equation}
\label{eq:emp-cvar-fractional}
\widehat{\mathrm{CVaR}}_\alpha(R)
 := \inf_{\eta\in\mathbb{R}}\widehat{\phi}_N(\eta)
=\frac{1}{t}\left(\sum_{i=N-m+1}^{N} R_{(i)}  +  sR_{(N-m)}\right).
\end{equation}
\end{proposition}



Proposition ~\ref{prop_emp-cvar-fractional}  shows that the empirical $\mathrm{CVaR}_\alpha$ can be expressed
as the average of the $m$  largest individual losses (when s=0), and as a convex combination of the Top-$m$ and a fractional contribution from $R_{(N-m)}$ (when $s>0$). 

\paragraph{}
Considering $N$ residual samples, let $t=(1-\alpha)N$, $m=\lfloor t\rfloor$, and $s=t-m\in[0,1)$, the
proposition~\ref{prop_emp-cvar-fractional} shows that the empirical CVaR always depends only on the \emph{tail set} of worst residuals:
\begin{equation}
\widehat{\mathrm{CVaR}}_\alpha(R)
= \frac{1}{t}\!\left(\sum_{i=N-m+1}^{N} R_{(i)}  +  sR_{(N-m)}\right).
\label{eq:emp-cvar-fractional-tail}
\end{equation}
If $s=0$ (i.e. $t$ is equal to is floor $m$), this reduces to a \emph{Top-$t$ average}:
\begin{equation}
\widehat{\mathrm{CVaR}}_\alpha(R)
= \frac{1}{t}\sum_{i\in\mathcal{I}_{\text{top-}t}} R_i,
\qquad
\mathcal{I}_{\text{top-}t} := \{ \text{indices of the $t$ largest residuals}\}.
\label{eq:emp-cvar-topk}
\end{equation}
When $s>0$, the value is a convex combination of the Top-$m$ and a fractional contribution from $R_{(N-m)}$, but the same \emph{tail weighting principle} holds: only the largest residuals matter.  This admits a natural \emph{adaptive weighting interpretation}. 
Define weights
\begin{equation}
w_i =
\begin{cases}
\dfrac{N}{t}, & i \in \mathcal{I}_{\text{tail-}t},\\[4pt]
0, & \text{otherwise},
\end{cases}
\label{eq:topk-weights}
\end{equation}
where $\mathcal{I}_{\text{tail-}t}$ is the tail set: the top-$t$ residuals if $s=0$, or the top-$m$ plus a fractional contribution from $R_{(N-m)}$ if $s>0$.
\begin{equation}
\widehat{\mathrm{CVaR}}_\alpha(R)
= \frac{1}{N}\sum_{i=1}^N w_i R_i.
\label{cvar_descirption}
\end{equation}
Thus CVaR can be viewed as a weighted empirical mean in which optimization effort is concentrated \emph{adaptively on the worst residuals}.
Unlike heuristic reweighting schemes commonly used in PINNs, the weights $(w_i)$ are uniquely determined by the coherent risk measure CVaR and thereby operationalize the probabilistic constraint
$P(|r(X;\theta)|\le \varepsilon)\ge \alpha$.

\subsection{RRaPINN Formulation}
\subsubsection{CVaR Hinge Penalty}

The empirical \emph{squared L1 exact penalty relaxation}  \citep{fletcher1985l1, smith1997penalty}, of the problem ~\eqref{pco_pinn_cvar} is written as:
\begin{equation}
\widehat{\mathcal{L}}_{\text{robust}}(\theta,\varepsilon)
= \widehat{\mathcal{L}}_{\text{PINN}}(\theta)
+ \lambda_p \Big(\widehat{\mathrm{CVaR}}_\alpha(R) - \varepsilon\Big)_+^2
+ \gamma_\varepsilon\varepsilon,
\label{eq:penalty}
\end{equation}
$R_i=|r(x_i;\theta)|$, and $(\cdot)_+=\max\{0,\cdot\}$. 
The term $\gamma_\varepsilon\varepsilon$ prevents the trivial solution $\varepsilon\to\infty$ and encodes a prior preference for tighter tolerances, $\lambda_p$ controls the strength of the penalty when the tail average exceeds $\varepsilon$.  Substituting the  value of $\mathrm{CVaR}_\alpha(R)$ given in  Eq.~\eqref{cvar_descirption}, into the equation ~\eqref  {eq:penalty}, yield the final RRAPINN objective
\begin{equation}
\widehat{\mathcal{L}}_{\text{robust}}(\theta,\varepsilon)
= \widehat{\mathcal{L}}_{\text{PINN}}(\theta)
+ \lambda_p \Bigg(\frac{1}{N}\sum_{i=1}^N w_i R_i - \varepsilon\Bigg)_+^2
+ \gamma_\varepsilon\varepsilon,
\label{eq:penalty-weighted} 
\end{equation}
where the weights $(w_i)$ as defined in ~\eqref{eq:topk-weights}, concentrate on the worst $t$ residuals (Top-$t$ average if $t\in\mathbb{N}$, fractional tail if $t\notin\mathbb{N}$).
This formulation provides a principled, data-driven reweighting scheme derived from a coherent risk measure. The final loss \eqref{eq:penalty-weighted} exhibits the weighting system explicitly, which is now a residual risk aware mean that shifts training focus onto the most problematic regions of the PDE domain in each batch.

\subsubsection{RRaPINN-Based Weighted Mean Squared (RRaPINN-WMS): CVaR Mean Excess Surrogate.}

\begin{proposition}[Proof \footnote{The proof of the proposition ~\ref{cvar_msq_prop} can be found the Appendix ~\ref{cvar_consistency_msq}}]
\label{cvar_msq_prop}
Fix $\alpha\in(0,1)$ and  Let $(R_i)_{i=1}^N$ be nonnegative residual magnitudes, let $t:= (1-\alpha)N $ be an integer (equal to its floor and ceiling).
Considering $\varepsilon\ge 0$,  the weights  $w_i=\begin{cases}
\dfrac{N}{t}, & i \in \mathcal{I}_{\text{tail-}t},\\[4pt]
0, & \text{otherwise},
\end{cases}$  satisfy
\begin{equation}
\frac{1}{N}\sum_{i=1}^N w_i = 1
\quad\text{(equivalently, }\sum_i w_i = N\text{).}
\label{eq:sumwN_}
\end{equation}
 and defining the mean-squares robust penalty as
\(
\mathcal{P}_{\mathrm{ms}}(R;\varepsilon,w)
 := 
\frac{1}{N}\sum_{i=1}^N w_i\big(R_i-\varepsilon\big)_{+}^{2},\)
 the following inequality hold:
\begin{equation}
\mathcal{P}_{\mathrm{ms}}(R;\varepsilon,w)
 \ge 
\big(\widehat{\mathrm{CVaR}}_{\alpha}(R)-\varepsilon\big)_+^{2} .
\label{cvar_consistency_msq_prop}
\end{equation}

\end{proposition}

\paragraph{Empirical RRaPINN-WMS Top-k/ Mean Excess Penalty }
Under the  conditions given in eq.~\eqref{cvar_consistency_msq_prop},  we can formulate the RRA weighted mean squared surrogate as follows:

\begin{equation}
\widehat{\mathcal{L}}_{\text{surrogate}}(\theta,\varepsilon)
= \widehat{\mathcal{L}}_{\text{PINN}}(\theta)
+ \lambda_p \cdot\frac{1}{N}\sum_{i=1}^N w_i\big(R_i-\varepsilon\big)_{+}^{2}
+ \gamma_\varepsilon\varepsilon,
\label{eq:robust-loss-topk}
\end{equation}
which preserves the CVaR weighting principle while producing stronger gradient signals in the presence of rare but extreme misfits.

\subsection{Adaptive Gradient-Free Tail Threshold \texorpdfstring{$\varepsilon$}{epsilon}}
Treating $\varepsilon$ as a learnable parameter is fragile because  the optimizer can reduce the hinge penalty by \emph{inflating} $\varepsilon$ instead of shrinking tail residuals \footnote{Theoretical justification in Appendix~\ref{subsec:grads}}. We therefore \emph{do not backpropagate} through $\varepsilon$ and update it from a detached tail statistic. 
We then move $\varepsilon$ toward a \emph{margin-shifted} gradient-free empirical CVaR using the following formula $t=(1-\gamma)\widehat{\mathrm{CVaR}}_\alpha(\,|r_\theta|\,)$, with $\beta=0.95$
using  an EMA that smooths \emph{upward} changes and a cap that makes \emph{downward} changes immediate:
$\varepsilon \leftarrow \min\!\bigl(\,\beta\,\varepsilon + (1-\beta)\,t, t\,\bigr).$
With  this strategy, no gradients flow through $\varepsilon$, so there is no incentive to inflate it. The margin keeps a positive gap so the tail penalty remains active near convergence. The EMA used here reduces sampling noise while  the cap prevents overshoot(K-averaging can be added for extra smoothing, but was not required in our runs.) 


\subsection{Two-Phase Training with Dynamic Balancing}
\label{subsec_tail-control}

We use a two-phase loss. \emph{Warmup} (first $K$ epochs; $K{=}1000$ unless stated) minimizes
$\mathcal{L}_{\text{base}}+\mathcal{L}_{\text{BC}}+\mathcal{L}_{\text{IC}}$.
\emph{Tail-control} (epochs $>K$) adds a remove $\mathcal{L}_{\text{base}}$ and  risk term
$\mathcal{L}_{\text{pen}}=\lambda_p\,\mathcal{L}_{\text{core}}$ where
$\mathcal{L}_{\text{core}}$ is either CVaR–hinge
$([\mathrm{CVaR}_\alpha(|r|)-\varepsilon]_+)^2$ or mean–excess
$\mathbb{E}([ |r|-\varepsilon]_+^2)$. To match scales we set
\[\lambda_p \propto \frac{S_b}{S_p}, \, \text{where}
\quad
S_b=\mathrm{EMA}_{\eta_{EMA}=0.9}[\det(\mathcal{L}_{\text{base}})], S_p=\mathrm{EMA}_{\eta_{EMA}=0.9}[\det(\mathcal{L}_{\text{core}})].
\]
Balancing is optional, however it  auto-balances bulk vs tail gradients. For \emph{stiff} problems we optionally invert the schedule (tail-first, then bulk).
\footnote{ More details about the algorithm are provided in the appendix \ref{algo_details}}
\subsection{Benchmarks and Error Metrics}
\subsubsection{Benchmark Models}
We respectively denote our models   RRaPINN and its variant RRaPINN-WMS as RRa and RRaWMS and  compared  them to
Convolutional Weighting PINN (CWP)\citep{si2025convolutionweightingmethodphysicsinformedneural}, Residual-Based Attention PINN (RBAPINN) \citep{ANAGNOSTOPOULOS2024116805} and PINN Baseline (Base) \citep{raissi2019pinns}, using the metrics defined in the next subsection. All benchmark models were trained using  their default parameters.
\subsubsection{Metrics}
\paragraph{Relative $\mathbf{L^2}$ and $L^\infty$ errors.}
Given a reference (ground-truth) solution $u^\star$ and a prediction $\hat u$ on a spatial domain
$\Omega\subset\mathbb{R}^d$ (and optionally time $t\in(0,T]$), we report
\begin{align}
\mathrm{Rel}L^2(u) =
\frac{\| \hat u - u^\star \|_{L^2(\Omega)}}{\|u^\star\|_{L^2(\Omega)}},
\qquad
L^\infty(u) = \| \hat u - u^\star \|_{L^\infty(\Omega)}.
\end{align}
In practice, we approximate these by sampling a finite set $\{x_j\}_{j=1}^N\subset\Omega$
(or space-time points $(x_j,t_j)$) and use the Monte Carlo approach defined as follows:
\begin{align}
\mathrm{Rel}L^2(u)  \approx 
\frac{\Big( \sum_{j=1}^N \lvert \hat u_j - u^\star_j\rvert^2 \Big)^{\!1/2}}
     {\Big( \sum_{j=1}^N \lvert u^\star_j\rvert^2 \Big)^{\!1/2}},
\qquad
L^\infty(u)  \approx  \max_{1\le j\le N} \lvert \hat u_j - u^\star_j\rvert.
\end{align}
\paragraph{$\alpha-th$-percentile:}
The $\alpha-percentile$ ($Q_{\alpha})$ of absolute point-wise  residuals  was also 
determnined for $\alpha=0.95$.

\noindent\textbf{ Empirical Tail Plot (1-CDF) plot.}
Let $r$ denote the residual magnitude and let $F(x)=\Pr(|r|\le x)$ be its cumulative distribution function (CDF). 
The plotted quantity is the \emph{survival} (complementary CDF)
\[
S(x)=1-F(x)=\Pr(|r|>x),
\]
shown on a logarithmic $y$-axis. At any fixed threshold $x$, a \emph{lower} curve means a smaller fraction of the domain has residuals larger than $x$ (i.e., better tail behavior). Two useful readouts are:
(i) the crossing at $S(x)=0.05$, which yields the $95^{\mathrm{th}}$ percentile $Q_{95}$ (since $5\%$ of points exceed $x$), and 
(ii) the rightmost extent of a curve, which approximates $L_\infty=\max |r|$.
Moreover, the area under $S(x)$ over the worst $(1-\alpha)$ mass is directly related to $\mathrm{CVaR}_\alpha$ (the average residual in the upper tail). 

\section{Experiments}
\label{experiments}
In all the experiments, we used the Adam optimizer \citep{kingma2017adammethodstochasticoptimization}; to improve convergence, we employ a cosine annealing learning rate schedule with minimum value $\eta_{\min}=10^{-5}$. 
Formally, the learning rate at epoch $t$ is given by
\begin{equation}
    \eta_{t} = \eta_{\min} + \tfrac{1}{2} \left(\eta_{0} - \eta_{\min}\right)\left(1+\cos\left(\frac{T_{\text{cur}}}{T_{\max}}\pi\right)\right),
\end{equation}
where $T_{\text{cur}}$ is the current epoch and $T_{\max}$ is the maximum number of epochs.  The initial learning rate $\eta_{0}$ was problem-specific.  For the RRaPINN-based models, unless stated otherwise, we use $\alpha=0.95$ and $\lambda_p=0.03$; for RRa-based model, gradients are clipped to a maximum norm of $5.0$ for stability.





\subsection{1D Heat Equation}
We assess our methods on  the 1D heat equation with high-frequency dynamics \citep{fernandez2013strong,si2025convolutionweightingmethodphysicsinformedneural}:
\begin{figure}[H]
\centering
\begin{minipage}{0.45\textwidth}
    \begin{equation*}
        \frac{\partial u}{\partial t}(t,x) 
        = \alpha \frac{\partial^2 u}{\partial x^2}(t,x),
        \quad (t,x) \in (0,T] \times (0,1),
    \end{equation*}
    we consider T=1 and $\alpha=\dfrac{1}{400\cdot \pi^2}$, with initial and boundary conditions
    \begin{align*}
        u(t,0) &= u(t,1) = 0, \qquad t \geq 0, \\
        u(0,x) &= \sin(20 \pi x), \qquad x \in [0,1].
    \end{align*}
    \label{eq:heat_eq_side}
    To guarantee satisfaction of the initial and boundary conditions, we enforce the neural network output in the form
\begin{equation}
    \hat{u}(t,x) = t  x  (1 - x)  \hat{u}_{NN}(t,x) + \sin(20 \pi x),
\end{equation}
where $\hat{u}_{NN}(t,x)$ is the raw output of the neural network.
\end{minipage}
\hfill
\begin{minipage}{0.5\textwidth}
    \centering
    \includegraphics[width=\linewidth]{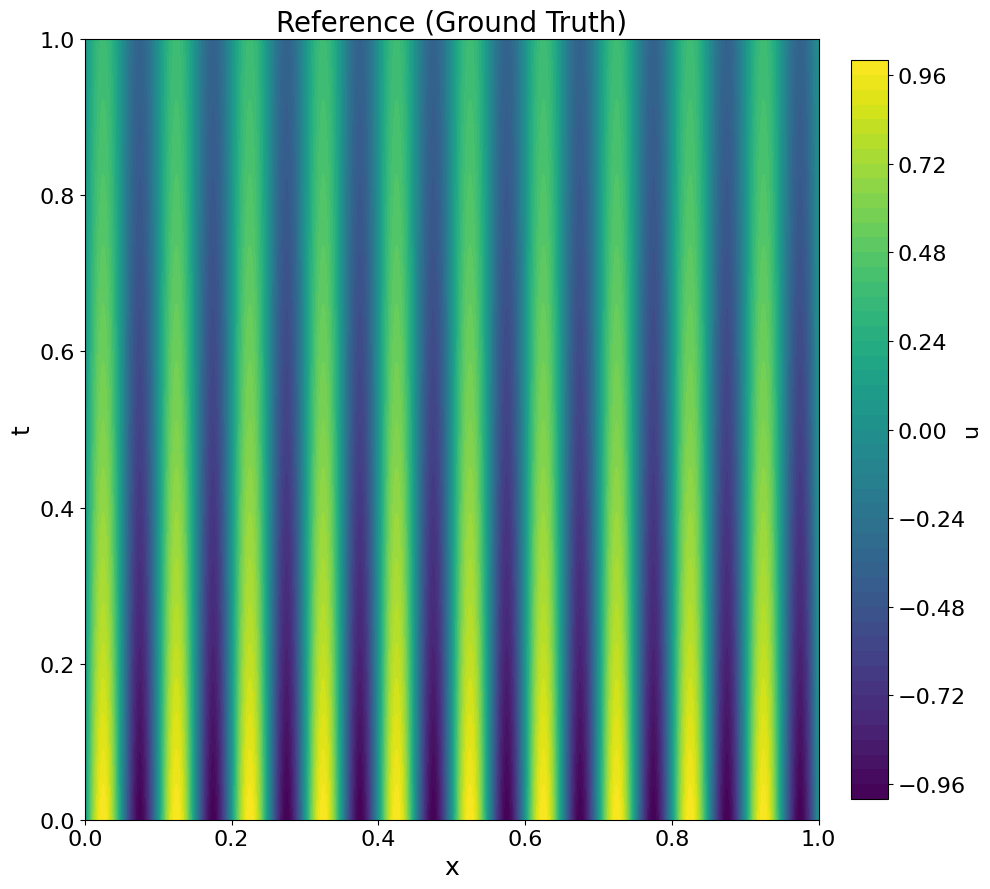} 
    \captionof{figure}{Reference solution for the 1D heat equation.}
    \label{fig_heat_true}
\end{minipage}%
\end{figure}


\paragraph{Neural network architecture and traininhg details.}The approximation $\hat{u}_{NN}$ is produced by a fully connected feedforward neural network with \emph{4 hidden layers of width 80}, each equipped with emph{hyperbolic tangent} activation functions. All weights are initialized using Xavier initialization and biases are set to zero. For training, we adopt the emph{Adam optimizer} with a given learning rate $\eta_0 = 9\times 10^{-3}$ over $15,0000$ iterations. We used $N_{tr}=5000$ interior point for the  training and use another $N_{te}=5000$  and initialization to foster diversity between train and test interior points). The average runtime per iteration was about $0.035\,s$ for RRa CVaR hinge and $0.04\,s$ for the RRa mean excess.



\begin{figure}[H]
\centering
\begin{minipage}{0.55\textwidth}
    \centering
    \includegraphics[width=\linewidth]{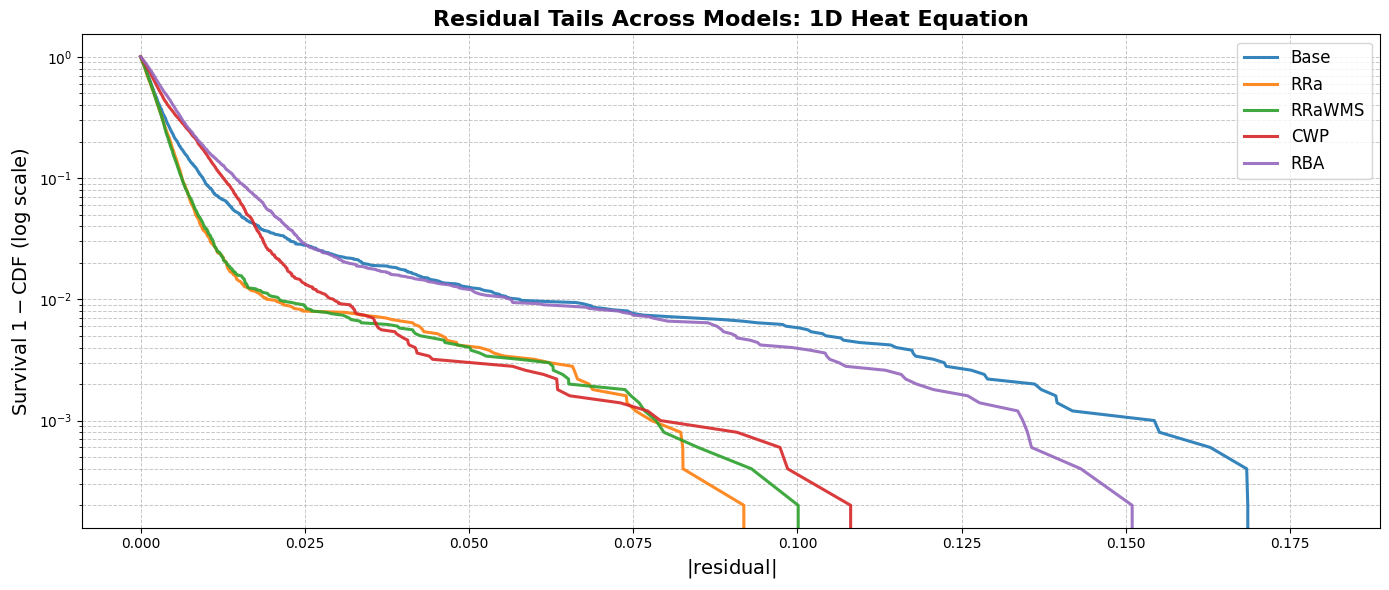} 
    \caption{Tail Plot for the 1D heat equation}
    \label{fig_heat_cdf}
\end{minipage}%
\hfill
\begin{minipage}{0.40\textwidth}
    \centering
    \begin{tabular}{lcccc}
    \toprule
   Method   &$\boldsymbol{L^2_{\mathrm{rel}}}$ & $\boldsymbol{L_\infty}$ & $\boldsymbol{Q_{95}}$ \\
\midrule
Baseline & 0.003340 & 0.019804 & 0.001676 \\
RRa      & \textbf{0.001331} & 0.008581 &  0.000826 \\
RRaWMS   & 0.001490 & 0.009460 &  \textbf{0.000732} \\
CWP      & 0.001640 & \textbf{0.006712} &  0.001726 \\
RBA      & 0.003498 & 0.019458 &  0.002145 \\
\bottomrule
    \bottomrule
    \end{tabular}
    \captionof{table}{Performance comparison across methods for the 1D heat equation.}
    \label{metrics_1dheat}
\end{minipage}
\end{figure}
\noindent
Table~\ref{metrics_1dheat} compares the performance of different methods on the 1D heat equation. 
RRa achieves the lowest relative $L_2$ error and MSE, indicating the most accurate prediction on average. 
CWP shows the smallest $L_\infty$ norm, suggesting better control over the worst-case pointwise errors. 
Looking at the 0.95-quantile metric, RRaWMS attains the lowest value, reflecting superior performance in nearly all points while still ignoring extreme outliers. 
Both RBA and Baseline perform similarly, indicating that residual-based weighting does not provide significant benefits for this smooth 1D heat equation problem.
The tail plot in figure \ref{fig_heat_cdf}, reveals that the RRa-based methods provide the steepest decay especially RRaWMS, followed by CWP. RBA and and Baseline PINN shows heavier tails. This indicates  that RRa-based methods foster  better robustness and extreme residual values control.

\subsection{2D Poisson Equation}

The Poisson equation is a fundamental elliptic PDE that models steady-state diffusion or potential fields, such as electrostatic or gravitational potentials. In two dimensions, it captures spatial interactions governed by Laplacian operators over a bounded domain. Owing to its smooth, well-posed nature and availability of analytical or high-accuracy numerical solutions, the 2D Poisson equation is widely used as a benchmark in physics-informed neural networks (PINNs). It provides a controlled setting to assess how well PINNs approximate spatial gradients, enforce boundary conditions, and generalize across varying geometries \cite{suhendar2024mesh}.

\begin{figure}[h!]
\centering
\begin{minipage}{0.45\textwidth}
   On a bounded Lipschitz domain $\Omega\subset\mathbb{R}^2$, the Poisson problem reads
\begin{align}
-\Delta u(x,y) = f(x,y) \quad \text{in }\Omega,
\end{align}
with either Dirichlet or Neumann boundary data:
\begin{align}
u = g \quad\text{on }\partial\Omega
\qquad\text{or}\qquad
\partial_n u = h \quad\text{on }\partial\Omega,
\end{align}
where $\partial_n$ denotes the outward normal derivative. For $f(x,y) = 2\pi^2 \sin(\pi x)\sin(\pi y)$, a common solution instance on $\Omega=(0,1)^2$ is
\begin{align}
u^\star(x,y) = \sin(\pi x)\sin(\pi y).
\end{align}
    \label{2dpoisson_eq_side}
    we use  $g = u^\star|_{\partial\Omega}$.
\end{minipage}
\hfill
\begin{minipage}{0.5\textwidth}
    \centering
    \includegraphics[width=\linewidth]{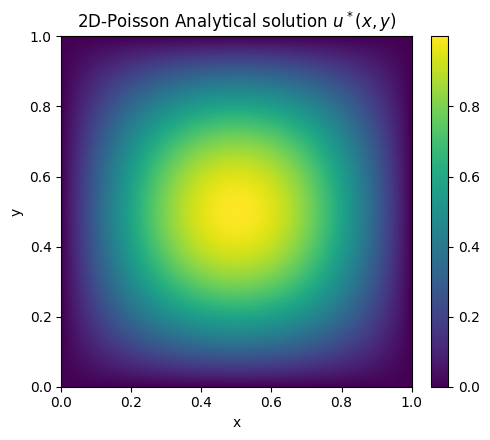} 
    \captionof{figure}{Exact Solution for the 2D Poisson equation.}
    \label{fig_2dpoisson_}
\end{minipage}%
\end{figure}

\paragraph{Neural network architecture and training details.}
We approximate $u(x,y)$ with a fully-connected multilayer perceptron (MLP)
 of \emph{depth 6} and \emph{width 64} with $\tanh$ activations.
The input is the spatial coordinate $(x,y)\in(0,1)^2$; the output is the scalar field value. We impose homogeneous Dirichlet data strongly by adding a boundary loss term. With gradient clipping at $5$
we train with Adam (learning rate $5\!\times\!10^{-3}$) and a cosine annealing schedule down to $10^{-5}$ over the full horizon. We use a warm-up of $1000$ epochs during which $L_{\mathrm{tail}}$ is inactive, and then enable it after warm up.
All experiments use single-precision on a single GPU. We use $10,000$ interior points and 200 boundary point.
For reporting, relative $L_2$ error is measured on a $101\times101$ grid against the manufactured solution $u^\star$.  The average runtime per iteration was about $0.010\,s$ for RRa CVaR hinge and $0.011\,s$ for the RRa mean excess.

\begin{figure}[H]
\centering
\begin{minipage}{0.55\textwidth}
    \centering
    \includegraphics[width=\linewidth]{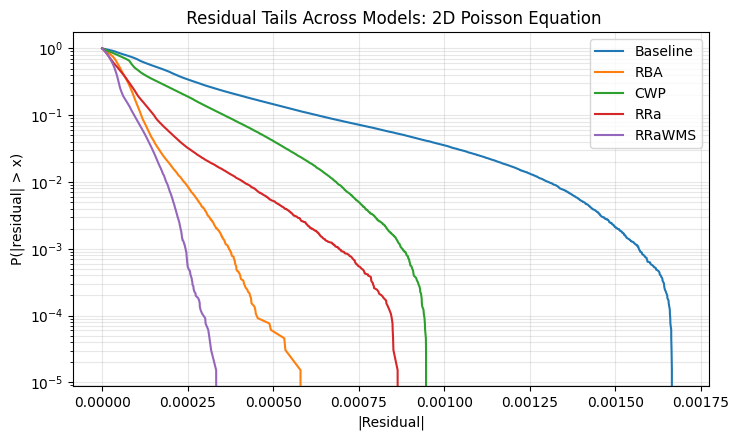} 
    \caption{Tail Plot for the 2D Poisson equation}
    \label{fig_2d_poisson_cdf}
\end{minipage}%
\hfill
\begin{minipage}{0.40\textwidth}
    \centering
   \begin{tabular}{lrrr}
\toprule
 Methods& $\boldsymbol{L^2_{\mathrm{rel}}}$ & $\boldsymbol{L_\infty}$ & $\boldsymbol{Q_{95}}$  \\
\midrule
Baseline & 0.000765 & 0.001689 & 0.000886 \\
RBA & 0.000156 & 0.000632 & 0.000141 \\
CWP & 0.000426 & 0.000947 & 0.000471 \\
RRa & 0.000212 & 0.000874 & 0.000207 \\
RRaWMS & \textbf{0.000118} & \textbf{0.000347} & \textbf{0.000127} \\
\bottomrule
\end{tabular}

    \captionof{table}{Performance comparison across methods for the 2D Poisson equation.}
\label{2d:poisson}
\end{minipage}
\end{figure}

Table ~\eqref{2d:poisson} shows that the best model across all metrics  is the RRaWMS, which is seconded by RBA and then RRa. CWP and Baseline PINN provide the worse metrics.   
Figure~\ref{fig_2d_poisson_cdf} shows survival curves \(S(a)=\Pr(|r|>a)\) for the 2D Poisson problem. \textsc{RRaWMS} (purple) lies farthest left with the steepest decay, indicating a uniformly thinner residual distribution from bulk to tail; this is consistent with its design that down-weights emerging large errors while preserving pressure on the bulk. \textsc{RBA} (orange) is next best and excels in the extreme tail (sharp drop near \(5\times10^{-4}\)), reflecting its targeted suppression of rare outliers. \textsc{RRa} (red) improves the mid-tail but cuts off later, while \textsc{CWP} (green) remains between \textsc{RRa} and the \textsc{Baseline} (blue), whose heavy tail reveals persistent hotspots. Therefore, it is worth saying that for a smooth elliptic PDE like 2D poisson, where residuals correlate closely with global \(H^1\)/\(L^2\) error and errors diffuse spatially, the leftward shift of the entire curve delivered by \textsc{RRaWMS} is particularly beneficial.

 \subsection{The Korteweg-de Vries (KdV) equation.}
\begin{figure}[ht!]
\centering
\begin{minipage}{0.45\textwidth}
  
We considered the one-dimensional form of the KdV equation  written as
\begin{align}
    \frac{\partial u}{\partial t}(x,t)  + 6u(x,t)\frac{\partial u}{\partial x}(x,t)  + \frac{\partial^3 u}{\partial x^3}(x,t) &= 0,\\ x\in \mathbb{R}\,,t\geq0, \, 
    \label{eq:kdv}
\end{align}
where $u(x,t)$ denotes the wave amplitude, $x$ the spatial coordinate, and $t$ the time variable. 
We consider the problem on a bounded periodic spatial domain $x \in [X_{\min}, X_{\max}]=[-10,10]$ and temporal domain $t \in [T_{\min}, T_{\max}]=[0,1]$, with periodic boundary conditions
\begin{align}
    u(X_{\min},t) = u(X_{\max},t),  
    u_x(X_{\min},t) = u_x(X_{\max},t),\\ \nonumber
    u_{xx}(X_{\min},t) = u_{xx}(X_{\max},t),
    \label{eq:kdv_bc}
\end{align}
and the analytical single-soliton initial condition, where $c>0$ (we considered $c=1$) denotes the soliton velocity and $x_0$ its initial position. 
\begin{equation}
    u(x,0) = \frac{c}{2}\,\mathrm{sech}^2\!\left( \frac{\sqrt{c}}{2}\,(x - x_0) \right),
    \label{eq:kdv_ic}
\end{equation}
\end{minipage}
\hfill
\begin{minipage}{0.45\textwidth}
    \centering
    \includegraphics[width=\linewidth]{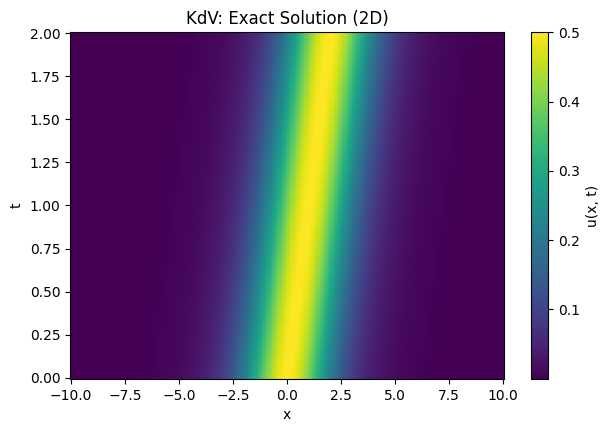} 
    \captionof{figure}{Exact Solution for the 1D Korteweg-de Vries Equation.}
    \label{fig_1d_kdv}
\end{minipage}%
\end{figure}

The Korteweg-de Vries (KdV) equation is a classical nonlinear partial differential equation that models the propagation of weakly nonlinear and weakly dispersive waves in shallow water channels \cite{korteweg1895change,raissi2019pinns}. 
It describes the interplay between nonlinear steepening and dispersive spreading, leading to the emergence of solitary wave solutions known as \emph{solitons} that retain their shape and velocity during propagation. This soliton travels without distortion at speed $c$, exemplifying the balance between nonlinearity and dispersion in Eq.~\eqref{eq:kdv}. 

From an optimization standpoint, KdV is nontrivial for PINNs because the third-order spatial derivative amplifies small numerical errors and sampling jitter, often producing heavy residual tails even when the bulk error is small.
These features nonlinearity, higher-order dispersion, and periodic constraints stress both the autodifferentiation stack and the loss geometry, motivating risk-aware formulations (e.g., tail-sensitive penalties) alongside standard mean-squared residual minimization.


\paragraph{Neural network architecture and training details.}
We approximate $u(x,y)$ with a fully-connected multilayer perceptron (MLP)
 of \emph{depth 4} and \emph{width 128} using \texttt{SiLU} activations.  Homogeneous Dirichlet data are imposed by adding a boundary loss term computed on $n_{\mathrm{bnd}}=512$ boundary samples per epoch; the interior residual is evaluated on $n_{\mathrm{int}}=10{,}000$ points. We train with AdamW (learning rate $1\times 10^{-2}$) and a cosine annealing schedule down to $10^{-5}$ over $10{,}000$ epochs, with gradient clipping at $1$. A \emph{warm-up} of $1{,}000$ epochs keeps the tail mechanism inactive; afterward, risk control is enabled with a broad tail split at $\alpha=0.95$. All runs use double precision on a single GPU. For reporting, the relative $L_2$ error is measured on a $201\times 201$ grid against the manufactured solution $u^\star$; residual-tail plots use the same grid.  The average runtime per iteration was about $0.074\,s$ for RRa CVaR hinge and $0.077\,s$ for the RRa mean excess.

\begin{figure}[ht!]
\centering
\begin{minipage}{0.45\textwidth}
    \centering
    \includegraphics[width=\linewidth]{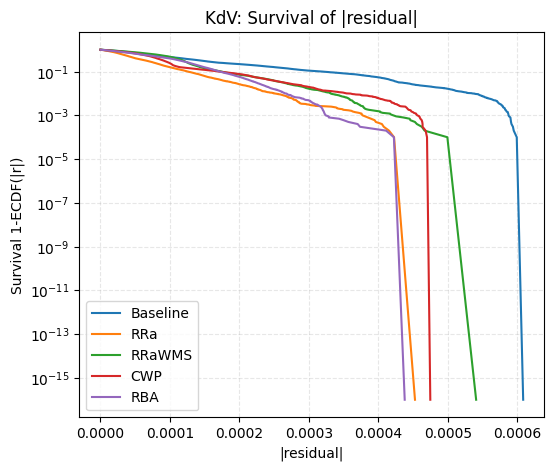} 
    \caption{Tail Plot for the 1D KdV  equation}
    \label{fig_1d_kdv_cdf}
\end{minipage}%
\hfill
\begin{minipage}{0.45\textwidth}
    \centering
\begin{tabular}{lrrr}
\toprule
\textbf{Model} & $\boldsymbol{L^2_{\mathrm{rel}}}$ & $\boldsymbol{L_\infty}$ & $\boldsymbol{Q_{95}}$ \\
\midrule
Baseline & 0.000297 & 0.000697 & 0.000057 \\
RRa      & 0.000264 & 0.000645 & 0.000058 \\
RRaWMS   & \textbf{0.000239} & 0.000613 & \textbf{0.000048} \\
CWP      & 0.000276 & 0.000586 & 0.000072 \\
RBA      & 0.000273&\textbf{0.000522}&0.000069\\
\bottomrule

\end{tabular}
    \captionof{table}{KdV PINN Performancees}
\label{1d_kdv_ccd}
\end{minipage}
\end{figure}
Table ~\eqref{1d_kdv_ccd}  shows that for the periodic, dispersive KdV dynamics (where errors in $u_{xxx}$ can diffuse and contaminate the field), RRaWMS attains the lowest $L^2_{\mathrm{rel}}$  and best $Q_{95}$, indicating uniformly small errors across space and tighter residual tails, desirable when third-derivative terms propagate local inaccuracies. RBA delivers the best $L_\infty$, meaning peaks/crests are clipped most tightly, but its higher $L^2_{\mathrm{rel}}$ and $Q_{95}$ suggest more localized hotspots elsewhere. CWP offers a balanced compromise (good $L_\infty$, mid $L^2_{\mathrm{rel}}$) but it provided $Q_{95}$), while  RRa provided the second best $L^2_{\mathrm{rel}}$ mid $Q_{95}$ and bad $L_\infty$. The Baseline provided the worse $L^2_{\mathrm{rel}}$ and  $L_\infty$ but the second best $Q_{95}$.

Figure~\ref{fig_1d_kdv_cdf} plots the complementary cumulative density function of absolute residual on a log scale: curves further left/down indicate better residual. 
 In the bulk range ($|r|\lesssim 3\times 10^{-4}$), \textsc{RRaWMS} stays lowest, implying fewer moderate violations and explaining its superior \(L^2_{\mathrm{rel}}\) and \(Q_{95}\). In the extreme tail (\(|r|\gtrsim 4\times 10^{-4}\)), \textsc{RBA} drops off most sharply, yielding the best $L_\infty$  by suppressing rare, large errors. The \textsc{Baseline} has the heaviest tail, indicating persistent large residuals. For KdV, the third-order dispersive term \(u_{xxx}\) magnifies curvature/high-frequency errors; thus, strong mid-tail control (as in \textsc{RRaWMS}) improves global waveform fidelity, while targeted suppression of extremes (as in \textsc{RBA}) prevents crest overshoots near steep gradients and dispersive fronts.

\subsection{1D Viscous Burgers' Equation}

The Burgers' equation serves as a canonical testbed for nonlinear advection-diffusion, illustrating the interplay between nonlinear steepening and viscous smoothing. In one dimension, the viscous Burgers' equation is often regarded as a stiff problem, since strong nonlinear advection can generate sharp gradients or shock-like structures that require fine temporal and spatial resolution to resolve, especially when viscosity is small \citep{li2015improved}.

\begin{figure}[H]
\centering
\begin{minipage}{0.5\textwidth}
    We solve the viscous Burgers PDE defined as follows.
    \begin{align}
        \partial_t u(x,t) + u(x,t)\partial_x u(x,t) 
        &= \nu\partial_{xx} u(x,t),
        \\  \nonumber \\
        (x,t)\in \Omega\times(0,1], \quad \Omega=(-1,1),
    \end{align}
    with kinematic viscosity $\nu>0$, initial condition
    \begin{align}
        u(x,0) &= -\sin(\pi x), \qquad x\in[-1,1],
    \end{align}
    and homogeneous Dirichlet boundary conditions
    \begin{align}
        u(-1,t) &= u(1,t) = 0, \qquad t\in[0,1].
    \end{align}
    \caption*{}
    \label{eq:burgers_side}
\end{minipage}
\hfill
\begin{minipage}{0.45\textwidth}
    \centering
    \includegraphics[width=\linewidth]{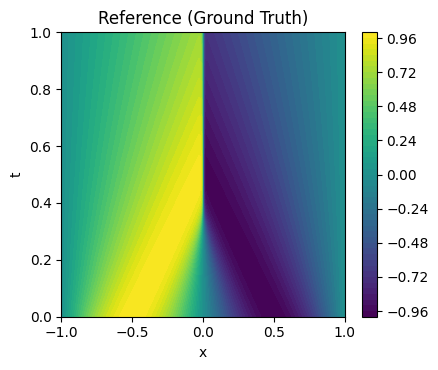} 
    \caption{Reference solution for the viscous Burgers equation.}
    \label{fig_burgers_prediction}
\end{minipage}%
\end{figure}
\begin{figure}[H] 
    \centering
    \begin{subfigure}{0.23\textwidth}
        \centering
        \includegraphics[width=\linewidth]{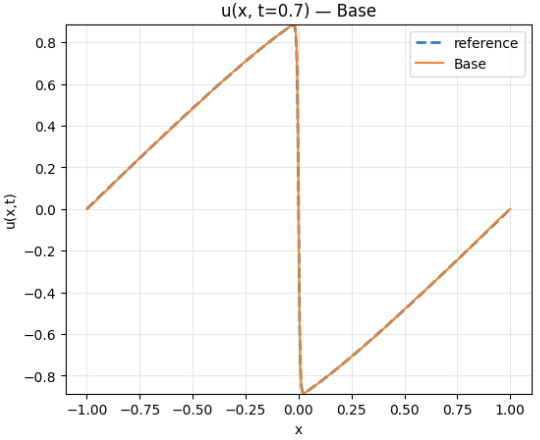}
        \caption{u at t=0.7}
        \label{burger_t0.7}
    \end{subfigure}
        \hfill
    \begin{subfigure}{0.23\textwidth}
        \centering
        \includegraphics[width=\linewidth]{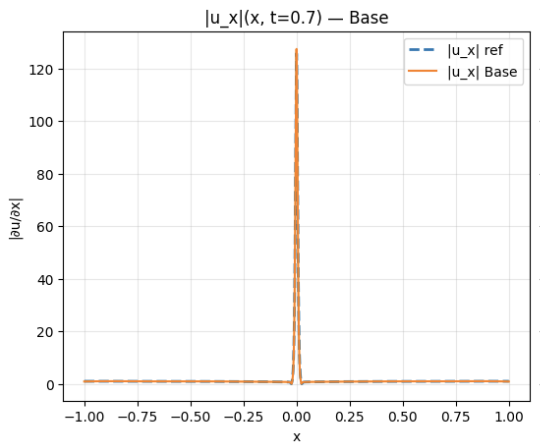}
        \caption{Gradient at t=0.7}
        \label{burger_t_0.7_grad}
    \end{subfigure}
    \hfill
    \begin{subfigure}{0.23\textwidth}
        \centering
        \includegraphics[width=\linewidth]{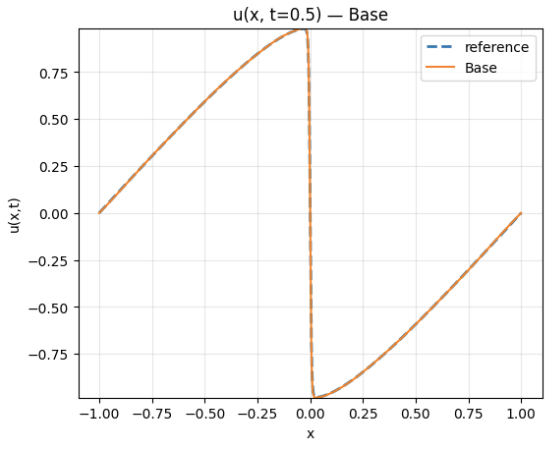}
        \caption{u at t=0.5}
        \label{burger_t0.5}
    \end{subfigure}
    \hfill
    \begin{subfigure}{0.23\textwidth}
        \centering
        \includegraphics[width=\linewidth]{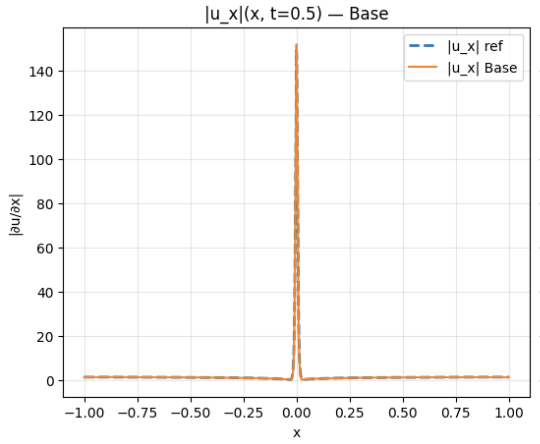}
        \caption{Gradient at t=0.5}
        \label{fig_burger_t0.5_grad}
    \end{subfigure}
    \caption{Predicted vs Actual 1D solution and gradients at fixed times.}
    \label{fig_both_images}
\end{figure}

We adopt the standard reference viscosity parameter 
$\nu = 0.01/\pi$.  The initial and boundary conditions were enforced in hard form, i.e., 
the network output was wrapped in a trial solution of the form
\begin{equation}
    U_{\text{Enf}}(x,t) = t(1 - x^2)  U_{\text{NN}}(x,t) - \sin(\pi x),
\end{equation}
which guarantees satisfaction of the prescribed constraints by construction.  
Figures~\eqref{fig_both_images} present the high resolution numerical solutions and their spatial gradients at $t=0.5$ and $t=0.7$. A pronounced gradient is clearly observed around $x=0$, highlighting the formation of a steep transition layer. This steepening is consistent with the characteristic dynamics of the viscous Burgers equation, demonstrating that the model accurately captures regions of rapid variation and sharp changes in the solution profile.
\paragraph{Neural network architecture and traininhg details.} For this set of experiments, the model was trained for 
\emph{20,000 epochs} using a fully connected feed-forward neural network with 
\emph{7 hidden layers}, each consisting of \emph{20 neurons}. The 
hyperbolic tangent ($\tanh$) activation function was employed 
throughout the network, with the final activation also set to $\tanh$.   Optimization was performed using the Adam optimizer with a learning 
rate of $\eta_0 = 5\times 10^{-3}$. To improve convergence, a  cosine learning rate scheduler was applied. The number of residual interior  points was fixed at \emph{10,000}. 
For evaluation, at each training iteration we sampled a test set of 
90,000 points (with new random initialization at every evaluation). The 
relative $L_2$ error was computed and recorded every 
500 epochs. The average runtime per iteration was about $0.021\,s$ for RRa CVaR hinge and $0.03\,s$ for the RRa mean excess.


\begin{figure}[H]
    \centering
    \includegraphics[width=1.1\linewidth]{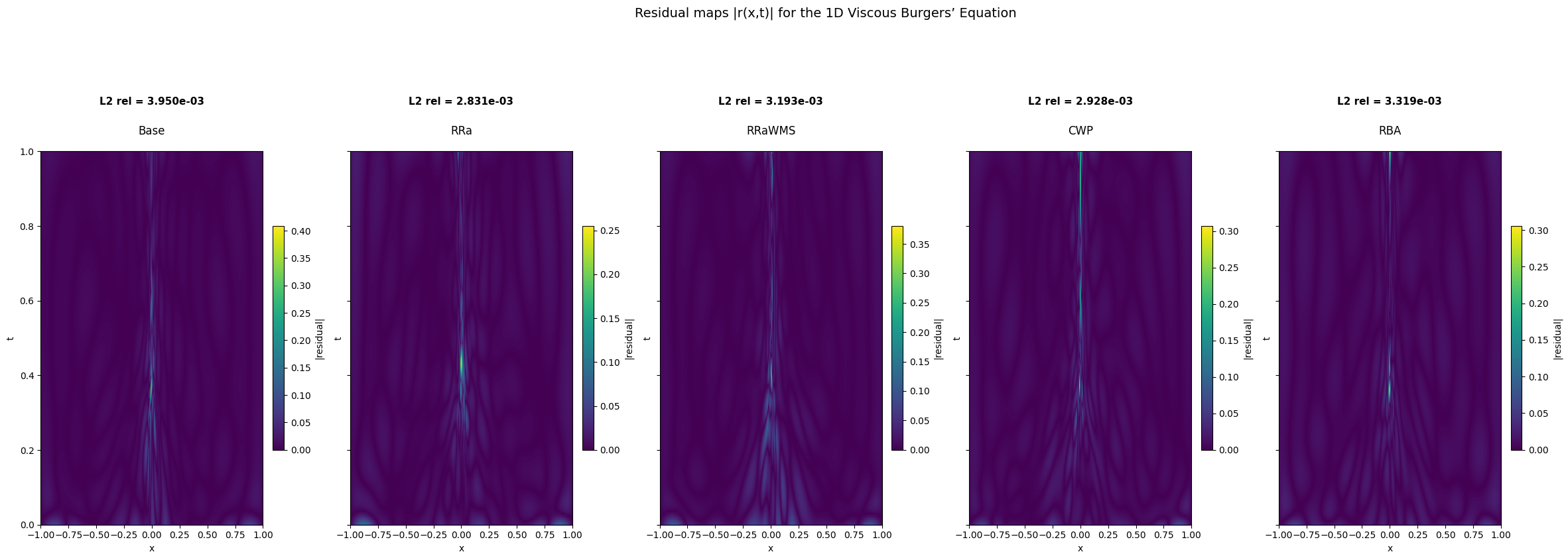}
    \caption{ Pointwise L2 Relative Error Plot for the 1D Burger Equation}
    \label{fig_placeholder_plot}
\end{figure}
Figures ~\eqref{fig_placeholder_plot} show that all the model predictions grow unexpectedly around $x=0$ and $t\in [0.4,1]$, however, RRA display the most controlled performances among all models, as well as the lowest L2 relative error overall which is followed by CWP, then RBAPINN, RRAWMS and the poorer result was delivered by the PINN Baseline.



\begin{figure}[H]
\centering
\begin{minipage}{0.55\textwidth}
       \centering
    \includegraphics[width=0.85\linewidth]{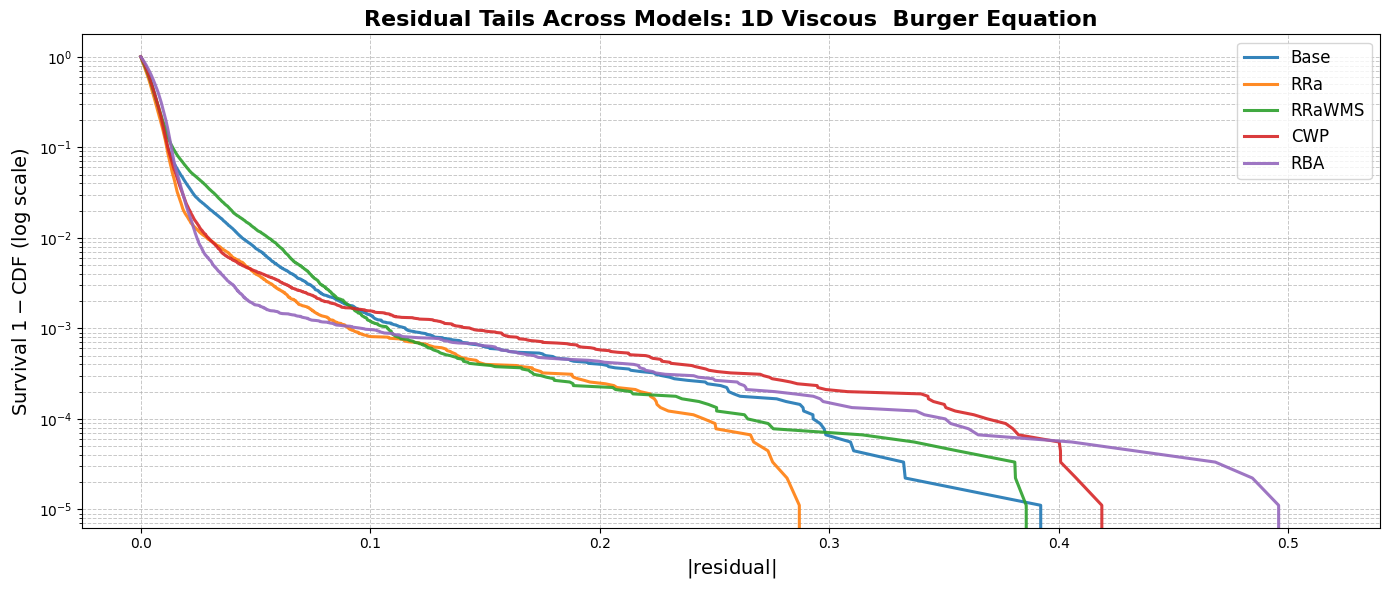}
    \caption{Tail Plot of Residuals (Residual vs Cummulative Proportion)}
    \label{fig_survival}
\end{minipage}%
\hfill
\begin{minipage}{0.40\textwidth}
    \centering
 \resizebox{1.1\linewidth}{!}{%
 \begin{tabular}{l *{3}{r} *{3}{r}}
\toprule
& \multicolumn{3}{c}{$t=0.5$} & \multicolumn{3}{c}{$t=0.7$} \\
\cmidrule(lr){2-4}\cmidrule(lr){5-7}
Method & $\boldsymbol{L^2_{\mathrm{rel}}}$ & $\boldsymbol{L_\infty}$ & $\boldsymbol{Q_{95}}$  & $\boldsymbol{L^2_{\mathrm{rel}}}$ & $\boldsymbol{L_\infty}$ & $\boldsymbol{Q_{95}}$ \\
\midrule
Base    & 0.00990 & 0.03940 & 0.02786 & 0.01041 & 0.03598 & 0.02727 \\
RRa     & 0.00692 & 0.02581 & \textbf{0.01707} & 0.00616 & 0.01942 & \textbf{0.01440} \\
RRaWMS  & 0.00953 & 0.03775 & 0.02730 & \textbf{0.00596} & \textbf{0.01753} & 0.01469 \\
CWP     & 0.00799 & 0.03083 & 0.01957 & 0.00732 & 0.02421 & 0.01578 \\
RBA     & \textbf{0.00665} & \textbf{0.02371} & 0.01821 & 0.00677 & 0.02136 & 0.01535 \\
    \bottomrule
    \end{tabular}
    }
    \caption{Comparison of methods on error metrics at the shock region for $t\!=\!0.5$ and $t\!=\!0.7$ (lower is better).}
\label{tab_rrapinn_results_both}
\end{minipage}
\end{figure}

Table~\eqref{tab_rrapinn_results_both} shows a clear split: RBA leads at the earlier time ($t=0.5$) on both global accuracy ($L_{2,\mathrm{rel}}$) and worst-case error ($L_\infty$), while RRaWMS takes the lead later ($t=0.7$) on the same metrics. RRa is the most consistent at controlling the error tail (lower $Q_{95}$) across both times. CWP performs solidly, but remains in the middle group overall.
Figure ~\eqref{fig_survival} reports the survival function of the absolute residuals on a log scale, highlighting the tail behavior of the error distributions across methods. The RRa-based approaches (RRa and RRaWMS) exhibit the steepest decay, indicating a substantially reduced frequency of large residuals. In contrast, the Base and CWP models show heavier tails, while RBA further extends the error distribution, occasionally producing markedly larger residuals. These results suggest that the RRa-based methods not only improve average predictive accuracy but also enhance robustness by suppressing extreme errors, however, it may overlook some important spots ($t=0.5$) because of the random collocation points resampling.


\subsection{The 2D Discontinuous Poisson  Problem}
We aim to solve the 2D Poisson equation with constant conductivity ($k=1$) on the unit square, governed by a discontinuous right-hand side (RHS):
\vspace{-1em}
\begin{figure}[ht!]
\centering
\begin{minipage}{0.45\textwidth}
\begin{align*}
    - \Delta u &= f(x,y) \quad &&\text{in } \Omega = (0,1)^2 \\
    u &= 0 \quad &&\text{on } \partial\Omega
\end{align*}
The primary challenge stems from the forcing term $f(x,y)$, which is piecewise defined with a jump discontinuity at the interface $x = 0.5$:
\begin{equation*}
f(x,y) = 
\begin{cases} 
    2\pi^2 \sin(\pi x)\sin(\pi y) & \text{if } x < 0.5 \\
    -6\pi^2 \sin(\pi x)\sin(\pi y) & \text{if } x \geq 0.5
\end{cases}
\end{equation*}
\end{minipage}
\hfill
\begin{minipage}{0.4\textwidth}
    \centering
    \includegraphics[width=\linewidth]{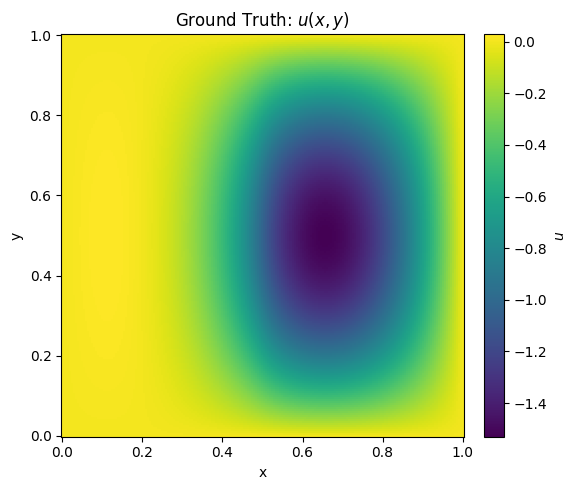} 
    \captionof{figure}{Numerical Solution for the 2D Discontinuous Poisson Equation.}
    \label{fig_2d_dis_pois}
\end{minipage}%
\end{figure}

Because the diffusion coefficient is constant, the physics dictates that the solution $u$ and its normal derivative $\partial_x u$ must be continuous across the interface. This forces the second derivative $\partial_{xx} u$ to have a jump discontinuity, creating a $C^1$  that standard $C^\infty$ sharp twist (kink) MLPs cannot represent without numerical oscillations (Gibbs phenomenon).
We augmented the standard loss with two explicit penalty terms to enforce the $C^1$ continuity at the interface $x=0.5$:   the $C^0$ continuity loss $L_{\text{iface}, u}=\mathbb{E}_{y \sim U[0,1]}[ (u(0.5^-, y) - u(0.5^+, y))^2 ]$ and   the $C^1$ continuity loss $L_{\text{iface}, u_x}=\mathbb{E}_{y \sim U[0,1]}[ (\partial_x u(0.5^-, y) - \partial_x u(0.5^+, y))^2 ]$.

\paragraph{Neural network architecture and training details.}
We approximate the solution $u(x,y)$ with a fully-connected multilayer perceptron (MLP) of \emph{depth 6} and \emph{width 64} using \texttt{Tanh} activations. The total loss is a composite of four distinct terms: the PDE residual $L_{\text{pde}}$, the Zero-Dirichlet boundary $L_{\text{bc}}$, and two explicit interface penalties, $L_{\text{iface}, u}$ ($C^0$ continuity) and $L_{\text{iface}, u_x}$ ($C^1$ continuity) at the $x=0.5$ discontinuity. The $L_{\text{pde}}$ is evaluated on $n_{\mathrm{int}}=4096$ interior points, resampled each epoch from a biased distribution that oversamples the interface. The $L_{\text{bc}}$ and $L_{\text{iface}}$ terms are evaluated on $n_{\mathrm{bc}}=2000$ and $n_{\mathrm{iface}}=2000$ points, respectively. We train with Adam (learning rate $1\times 10^{-3}$) and a cosine annealing schedule down to $10^{-5}$ over $15{,}000$ epochs, with gradient clipping at $1.0$. A \emph{warm-up} of $3{,}000$ epochs is used, during which the optimizer \textit{only} sees the $L_{\text{pde}}$ and $L_{\text{bc}}$ terms. After warmup, risk control is enabled: the interface losses ($L_{\text{iface}, u}$ and $L_{\text{iface}, u_x}$) are bundled with the CVaR penalty ($L_{\text{cvar}}$) at $\alpha=0.95$ to form a single $L_{\text{pen}}$ term. This $L_{\text{pen}}$ is then dynamically balanced against the $L_{\text{pde}}$ term using our adaptive weighting scheme. The $L_{\text{bc}}$ term remains active throughout with a static weight. All runs use double precision. For reporting, the metrics are  measured on an independent $201\times 201$ grid against a high-resolution finite difference solution $u^\star$. The average runtime per iteration was about $0.014\,s$ for RRa CVaR hinge and $0.016\,s$ for the RRa mean excess.


\begin{figure}[ht!]
\centering
\begin{minipage}{0.45\textwidth}
    \centering
    \includegraphics[width=\linewidth]{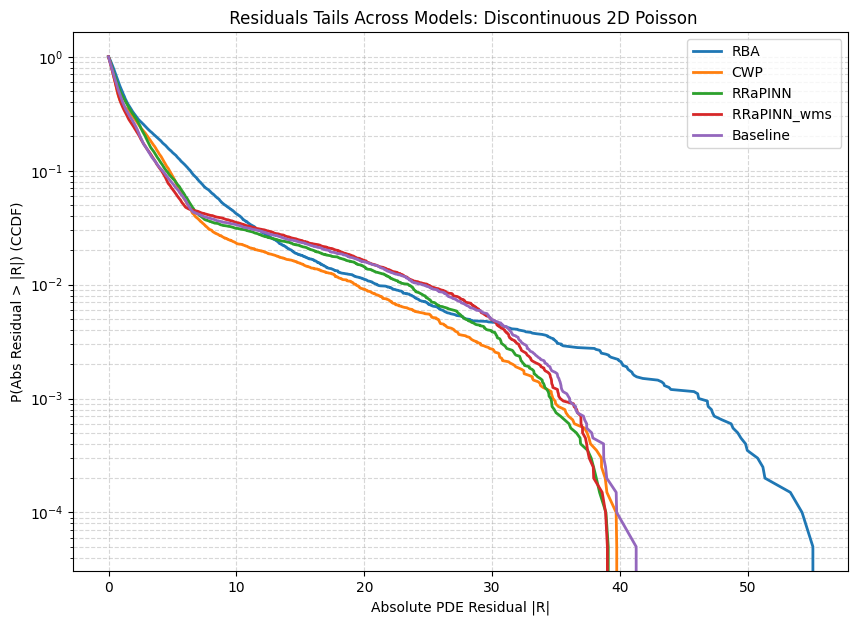} 
    \caption{Tail Plot for the 2D Discontinuous Poisson  equation}
    \label{fig_2d_dispoiss_cdf}
\end{minipage}%
\hfill
\begin{minipage}{0.45\textwidth}
    \centering
\centering
\begin{tabular}{lccc}
\hline
\textbf{Model} &   $\boldsymbol{L^2_{\mathrm{rel}}}$ & $\boldsymbol{L_\infty}$ & $\boldsymbol{Q_{95}}$ \\
\hline
RBA         & 0.09435 & 0.3331 & 8.901 \\
CWP         & 0.1030  & 0.3445 & 6.243 \\
RRa     & 0.08626 & \textbf{0.2997} & 6.340 \\
RRaWMS&\textbf{ 0.08305} & 0.3276 & \textbf{5.749 }\\
Baseline    & 0.1019  & 0.3442 & 6.283 \\
\hline
\end{tabular}
\caption{2D Discontinuous Poisson Performances}
\label{2d_dicont_poisson}
  
\label{2d_dicont_poisson_ccd}
\end{minipage}
\end{figure}
The CCDF plot (Figure ~\ref{fig_2d_dispoiss_cdf}) of absolute residuals  and table ~\ref{2d_dicont_poisson} for the discontinuous 2D Poisson case  show that the two risk-aware variants dominate the tail. In particular, RRaPINN-WMS (ME surrogate) achieves the lowest $P95$ residual (\(5.749\)), improving on the Baseline (\(6.283\)) and RBA (\(8.901\)) by roughly \(8.5\%\) and \(35\%\), respectively, and closely followed by RRaPINN (CVar hinge surrogate) (\(6.340\)) and CWP (\(6.243\)). The extreme right tail in the CCDF reveals a markedly heavier tail for \textbf{RBA}, with non-negligible mass persisting beyond residual levels \(r\!\approx\!40\)-\(55\), whereas the other methods decay around \(r\!\approx\!38\)-\(41\). This indicates that RBA can leave rare, large violations near the interface, while risk-aware training explicitly suppresses such extremes.
RRaPINN-WMS also delivers the best bulk error (L2 rel \(=\) \(0.08305\)), improving on the Baseline (\(0.1019\)); RRaPINN attains the lowest $L_\infty$ (\(0.2997\) vs Baseline \(0.3442\)), suggesting fewer localized spikes. Overall, the RRaPINN-based method reduce high-quantile residuals without sacrificing bulk accuracy, yielding a more reliable solution under discontinuities than  the Baseline PINN (Vanilla), RBA its variants CWP.

\subsection{Ablation study on the tail level $\alpha$}
We study how the tail level $\alpha$, for $\alpha\in\{0.50,0.75,0.85,0.95,0.99\}$ affects accuracy and residual control across three PDEs on $[0,1]$, using the 1D heat, 1D viscous Burgers, and 2D Poisson to provide  stable insights \footnote{Individual  plot are provided in the Appendix \ref{ablation_studies}}). The network is a fully-connected multilayer perceptron (MLP) with \emph{4 hidden layers} and \emph{80 neurons} per layer, 
employing the TanH activation function, with Adam optimizer learning rate $5\cdot10^{-3}$, boundary/initial weights similar as the above mentioned experiments, and training budget are held fixed. After a T-warmup (T=1000 epochs), we switch to a \emph{tail-only} objective with a mean-excess penalty and CVaR hinge ; the tail threshold $\varepsilon_t$ was updated each step from a detached CVaR estimate via an EMA with a small relative margin (no gradients flow through $\varepsilon_t$) and  initialized  at $\varepsilon_{t=0}=0.5$ for all problems. We trained our models  over 10,000 iterations for $\alpha\in\{0.50,0.75,0.85,0.95,0.99\}$,  and report test metrics on uniform grids (250$\times$250).  We compute the relative $L_2$ error, the  mean absolute residual, and  the final threshold $\varepsilon$; and \emph{average} of these metrics over the three PDEs (per-problem curves are provided in the appendix), which are presented in the figures ~\ref{fig_ablation_mean_excess}. This setup isolates the effect of $\alpha$ on the active tail set and the strength of the tail gradients while keeping all other factors constant.

\subsubsection{CVaR-hinge penalty}
\begin{figure}[ht!]
    \centering
    \includegraphics[width=0.65\linewidth]{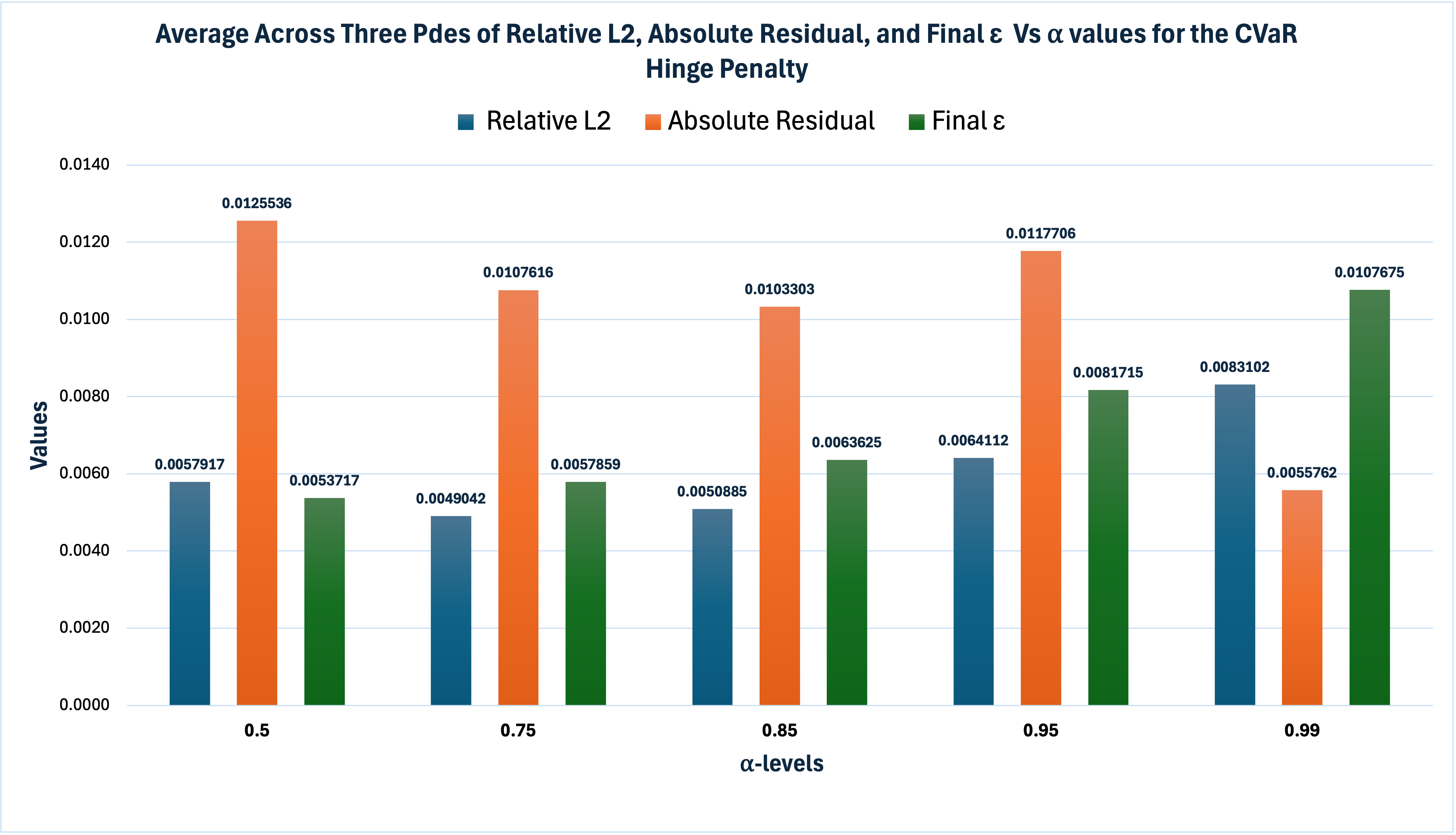}
    \caption{Ablation of the tail level $\alpha$ for the \emph{CVaR-hinge} penalty, averaged over the 1D heat, 1D viscous Burgers, and 2D Poisson problems. We report relative $L_2$, mean absolute residual, and the final threshold $\varepsilon_t$.}
    \label{fig_ablation_cvar_hinge}
\end{figure}

In Figure~\ref{fig_ablation_cvar_hinge}  metrics such as (Rel.$L_2$ and the mean absolute residual)  improve as $\alpha$ increases from $0.50$ to the mid range ($\alpha\!\approx\!0.75$-$0.85$), then worsen again for very large $\alpha$. However, the mean absolute residual takes its smallest value decreases again $\alpha=0.99$ while
the final $\varepsilon$ keeps on increasing as $\alpha$ increases.
This matches the behavior of the CVaR-hinge,
\(
P_{\text{hinge}}(\theta,\varepsilon)=\bigl[\mathrm{CVaR}_\alpha(R_\theta)-\varepsilon\bigr]_+^{\,2},
\)
with $\varepsilon$ updated by a detached EMA toward $(1-\textit{margin})\,\mathrm{CVaR}_\alpha$.
At low $\alpha$ the ``tail" is broad, gradients resemble a smoothed MSE and dilute attention to the worst regions.  
Mid-range $\alpha$ concentrates weight on the hardest samples while preserving enough coverage for stable, low-variance updates, hence the best average errors.  
As $\alpha$ grows further, $\mathrm{CVaR}_\alpha$ (and thus $\varepsilon_t$) rises, the active set shrinks, and training over-focuses on a few extremes; this helps worst-case control but hurts bulk metrics (relative $L_2$ and mean residual).

\paragraph{Practical takeaway.}
For CVaR-hinge, a moderate tail level is a better spot: $\alpha\!\in[0.75,0.85]$ typically balances tail minimization and bulk accuracy.  
Use a small relative margin (5-20\%) and an EMA $\beta\!\in[0.7,0.9]$; increase $\alpha$ (and possibly $\lambda_p$) only if the goal is stricter $L_\infty$ control.

\subsubsection{Mean-excess penalty} 
\begin{figure}[ht!]
    \centering
    \includegraphics[width=0.65\linewidth]{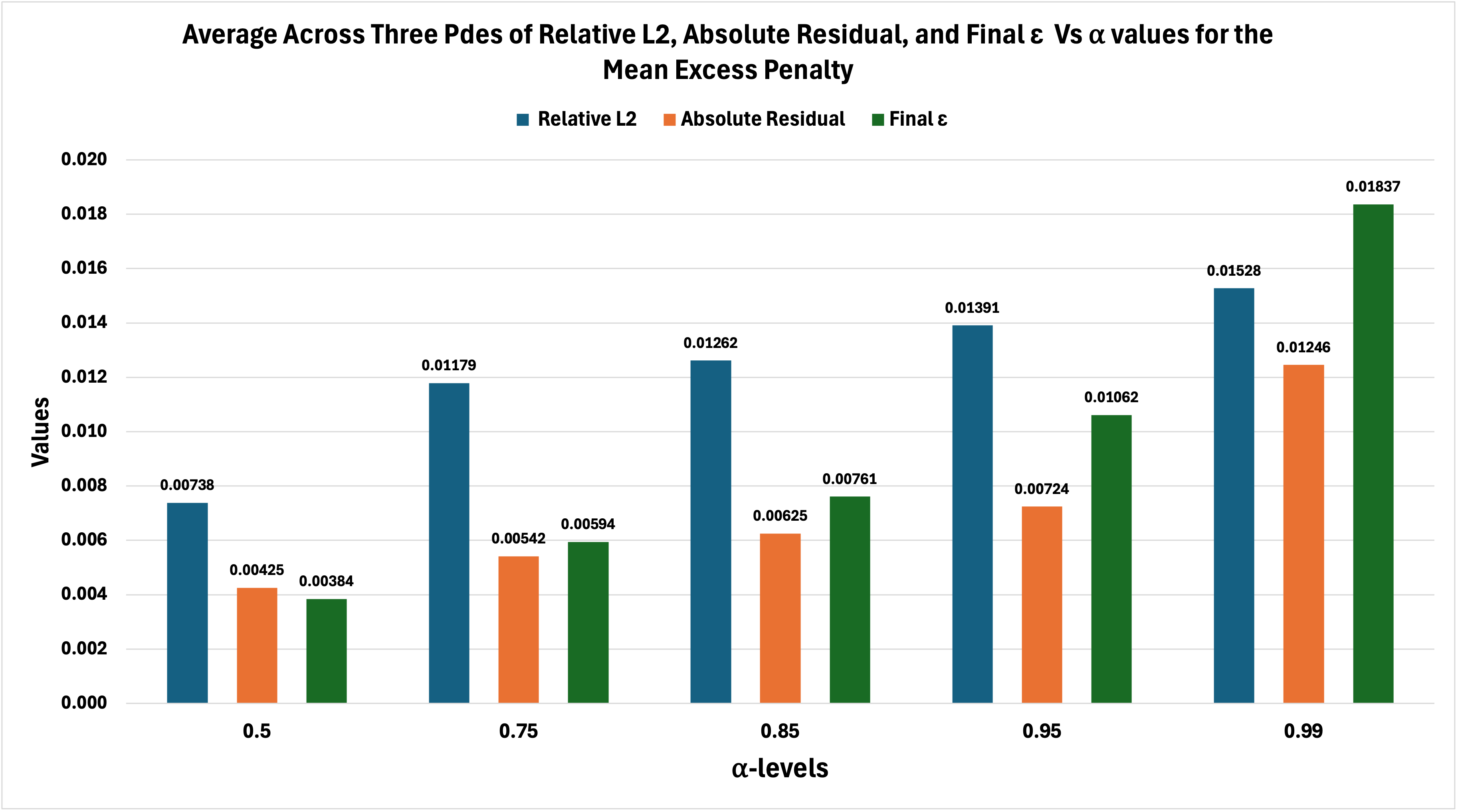}
    \caption{Ablation of the tail level $\alpha$ for the \emph{mean-excess} penalty, averaged over the 1D heat ($x,t$), 1D viscous Burgers ($x,t$), and 2D Poisson ($x,y$) problems. We report relative $L_2$, mean absolute residual, and the final threshold $\varepsilon_t$.}
    \label{fig_ablation_mean_excess}
\end{figure}

Figure~\ref{fig_ablation_mean_excess} shows a clear trend: as the tail level $\alpha$ increases (i.e., the tail becomes \emph{tighter}), the average relative $L_2$ error, the mean absolute residual, and the final threshold $\varepsilon_t$ all increase. This behavior is consistent with how the mean-excess penalty works. For smaller $\alpha$ the active set contains a larger portion of the domain, yielding lower-variance gradients and more benefit to the \emph{bulk} error. As $\alpha$ grows, the penalty concentrates on ever rarer extremes; the CVaR at level $\alpha$ rises, so the EMA-driven threshold $\varepsilon_t \approx (1-\textit{margin})\,\mathrm{CVaR}_\alpha$ also rises, and fewer points remain above the threshold. The result is stronger pressure on worst-case errors (useful for $L_\infty$ control) but less improvement on average metrics.

\paragraph{Practical takeaway.}
For mean-excess, moderate tails typically balance stability and coverage: $\alpha\!\in\![0.5,\,0.95]$ with a small relative margin (5-20\%) and EMA $\beta\!\in\![0.7,\,0.9]$ are recommended. Very large $\alpha$ focuses too narrowly, inflates $\varepsilon_t$, and can under-serve the bulk.

\section{Conclusion and Future Work}
In this work, we introduce the \emph{Residual Risk-Aware Physics-Informed Neural Network} (RRaPINN) framework. Our formulation unifies adaptive weighting in PINNs with distributionally robust optimization, yielding models that explicitly control high quantiles of the residual distribution rather than only the average discrepancy. We further derive a mean-excess (ME) surrogate for the CVaR penalty, which retains the same risk-sensitive interpretation while being easier to implement and tune. Classical weighting heuristics can be reinterpreted as special cases of our probabilistic view, and by leveraging CVaR-style risk measures we provide a direct bridge between risk-sensitive optimization and scientific machine learning. This offers a principled pathway toward PINNs that are not only accurate on average but whose worst-case residuals can be targeted and, in principle, bounded across the domain. 

Practically, RRaPINNs leave both the PDE formulation and the network architecture unchanged: they operate purely at the loss level, keeping the standard mean residual as a base term and augmenting it with a simple risk-aware tail penalty. In this sense, they can be viewed as a drop-in modification of the residual loss with a small, interpretable set of knobs (notably the reliability level~$\alpha$) that trades bulk accuracy (lower~$\alpha$) against tail control (higher~$\alpha$). Across 1D viscous Burgers, 1D heat, KdV, and Poisson equations (including a discontinuous forcing variant), RRaPINNs consistently improved tail behavior, reducing high residual quantiles and $L_\infty$ error while maintaining or improving bulk error relative to vanilla PINNs and strong baselines such as Residual-Based Attention, even in discontinuity regimes. In our experiments, the mean-excess (ME)  surrogate of the CVaR was numerically more stable and achieved stronger tail control than a direct CVaR hinge.

Despite these benefits, RRaPINNs retain several limitations that point to natural improvements. Because collocation points are resampled independently at each epoch, the method is effectively memoryless, so difficult regions can be under-visited and the tail objective may chase transient hotspots rather than persistently fixing the same defects. The current risk is also applied to a \emph{global} residual distribution, which can allow a single high-error subregion (e.g., a sharp interface) to dominate the tail budget and leave other parts of the domain under-corrected. Finally, risk-awareness is restricted to PDE residuals, while boundary, initial, and data-fidelity terms remain in standard MSE form, even when they drive the overall error. Future work could introduce lightweight memory mechanisms for hard points, adopt spatially localized risk budgets, and extend risk-aware training to BC/IC and data terms to better balance bulk accuracy and tail control.

Clarifying these questions would turn RRaPINNs from an empirically strong method into one with principled guaranties for both smooth and discontinuous PDEs.


\section{Acknowledgement}
This work was made possible by a grant from Carnegie Corporation (provided through the African Institute for Mathematical Sciences). I extend my sincere gratitude to my graduating institution, the University of KwaZulu-Natal (UKZN), for providing a rigorous academic environment and the opportunities that have shaped my personal and professional growth.


\newpage
 \bibliographystyle{plainnat}  
\bibliography{main}   
\newpage
\appendix
\section*{Appendix}
\label{appendix}



\subsection{Residual Risk Aware PINNs Formulation}
\label{lem:phi-subdiff}
We have recast PINN training as a \emph{risk-aware} program:
\[
\min_{\theta,\varepsilon>0}  \underbrace{\mathcal{L}_{\text{PINN}}(\theta)}_{\text{physics + boundary}}
\quad \text{s.t.}\quad
\underbrace{\mathrm{CVaR}_\alpha\big(|r(X;\theta)|\big)}_{\text{tail risk of residuals}}
\le \varepsilon,
\]
and provided its empirical, optimizable relaxation \eqref{eq:robust-loss-topk}, whose adaptive Top-$k$ weighting arises \emph{necessarily} from the convex CVaR formulation. 
This yields a principled, tail-robust PINN that targets reliability beyond average accuracy and stands in contrast to heuristic reweighting or sampling rules.

\subsection{RRaPINN Algorithm Details}
\label{algo_details}

\subsubsection{RRaPINN Algorithm Pseudocode}

The core of our method is a two-phase training process. After a standard MSE-based warmup, the algorithm transitions to a ``strict tail" phase. In this phase, the optimizer is driven only by the anchor (IC/BC) losses and a dynamic risk-averse penalty. Crucially, the weight of this penalty, $\lambda_p$, is computed dynamically using detached, exponentially-moving-average (EMA) scales of the base (MSE) loss and the core penalty, a mechanism inspired by risk-balancing adaptation.

\begin{algorithm}[H]
\caption{RRaPINN (CVaR-Hinge or Mean-Excess) with Stiff/Smooth Scheduling}
\label{alg:rrapinn}
\begin{algorithmic}[1]
\Require epochs $E$, warmup $W$, CVaR level $\alpha$, anchor weights $(\lambda_{ic},\lambda_{bc})$, base penalty seed $\lambda_{\mathrm{cfg}}$,
risk init $\varepsilon_{\mathrm{init}}$, EMA decays $(\beta_s,\beta_\varepsilon)$, margin $\eta$, bounds $(\lambda_{\min},\lambda_{\max})$, schedule $\in\{\texttt{SMOOTH\_FIRST},\texttt{STIFF\_FIRST}\}$
\State \textbf{Init:} $\varepsilon\gets\varepsilon_{\mathrm{init}}$, $\lambda_p\gets\lambda_{\mathrm{cfg}}$, $S_b\gets 1$, $S_p\gets 1$, optimizer $\mathcal{O}$
\For{$e=1,\dots,E$}
  \State Sample interior $X_{\mathrm{int}}$, IC $(X_{\mathrm{ic}},u_{\mathrm{ic}})$, BC $(X_L,u_L,X_R,u_R)$
  \State $L_{\mathrm{ic}}\gets \mathrm{MSE}\big(f_\theta(X_{\mathrm{ic}}),u_{\mathrm{ic}}\big)$;\quad
         $L_{\mathrm{bc}}\gets \tfrac12\big[\mathrm{MSE}(f_\theta(X_L),u_L)+\mathrm{MSE}(f_\theta(X_R),u_R)\big]$
  \State $r\gets \mathcal{R}(X_{\mathrm{int}};\theta)$;\quad $L_{\mathrm{base}}\gets \mathrm{mean}(r^2)$
 \State $L_{anchor} \leftarrow \lambda_{ic} L_{ic} + \lambda_{bc} L_{bc}$
  \State \textit{Tail stat \& threshold (short):}
         \quad $c\gets \mathrm{CVaR}_\alpha(|r|)$;\quad
         $\varepsilon \gets \min\!\big(\varepsilon,\ \mathrm{EMA}_{\beta_\varepsilon}((1-\eta)c,\ \varepsilon)\big)$

  \State \textbf{Penalty core:}
  \If{\texttt{HINGE}} $L_{\mathrm{core}}\gets \big[\max(0,\,c-\varepsilon)\big]^2$
  \Else \texttt{(ME)} $L_{\mathrm{core}}\gets \mathrm{mean}\!\big(\max(0,\,|r|-\varepsilon)^2\big)$
  \EndIf

  \State \textit{ Partial balancer (Optional):}
         \quad $S_b\!\gets\!\mathrm{EMA}_{\beta_s}(S_b,\ \mathrm{stopgrad}(L_{\mathrm{base}}))$;\ \
         $S_p\!\gets\!\mathrm{EMA}_{\beta_s}(S_p,\ \mathrm{stopgrad}(L_{\mathrm{core}}))$;\ \
         $\lambda_p\!\gets\!\mathrm{clip}\!\left(\lambda_{\mathrm{cfg}}\frac{S_b}{S_p+\delta},\,\lambda_{\min},\lambda_{\max}\right)$

  \State \textbf{Schedule \& loss:}
  \If{\texttt{SMOOTH\_FIRST}}
     \If{$e\le W$} $L\gets L_{\mathrm{base}}+\lambda_{ic}L_{\mathrm{ic}}+\lambda_{bc}L_{\mathrm{bc}}$
     \Else $L\gets L_{\mathrm{base}}+\lambda_p L_{\mathrm{core}}+L_{anchor}$
     \EndIf
  \Else \texttt{(STIFF\_FIRST)}
     \If{$e\le W$} $L\gets \lambda_p L_{\mathrm{core}}+L_{anchor}$
     \Else $L\gets L_{\mathrm{base}}+\lambda_p L_{\mathrm{core}}+\lambda_{ic}L_{\mathrm{ic}}+\lambda_{bc}L_{\mathrm{bc}}$
     \EndIf
  \EndIf

  \State \textbf{Update:} $\theta \gets \mathrm{Step}\big(\mathcal{O},\, \nabla_\theta L\big)$ with gradient clipping
\EndFor
\State \textbf{return} $\theta$
\end{algorithmic}

\vspace{0.6em}
\noindent\textbf{Note.} For \emph{stiff} problems, prefer \texttt{STIFF\_FIRST}: control the tail (penalty) during warmup, then emphasize bulk. For \emph{smooth} problems, prefer \texttt{SMOOTH\_FIRST}: warm up on the bulk, then add tail control.
\end{algorithm}

\subsection{From CVar to the Empirical CVaR via the Rockafellar-Uryasev Program}
\label{UR_Program}

\begin{remark}[Connection between CVaR and Average Top-$k$ Loss]
The empirical $\mathrm{CVaR}_\alpha$ formulation,
\begin{equation}
\label{eq:cvar_min_form}
\mathrm{CVaR}_\alpha(R)
=\min_{\lambda\in\mathbb{R}}
\left\{
\lambda+\frac{1}{(1-\alpha)N}
\sum_{i=1}^N [\,R_{i}-\lambda\,]_+
\right\},
\end{equation}
is mathematically equivalent to the \emph{Average Top-$k$ (AT$_k$) loss} introduced by \citet{NIPS2017_6c524f9d},
\begin{equation}
\label{eq:atk_form}
\mathcal{L}_{\mathrm{AT}k}
=\min_{\lambda\ge0}
\left\{
\frac{1}{N}\sum_{i=1}^N [\,R_{i}-\lambda\,]_+
+\frac{k}{N}\lambda
\right\}.
\end{equation}
Identifying $\alpha=1-\tfrac{k}{N}$ shows that minimizing CVaR at level $\alpha$ is equivalent to averaging 
the largest $k$ individual losses, that is, the top-$k$ tail of the empirical loss distribution as defined in \citep{NIPS2017_6c524f9d}.
This provides a direct bridge between risk-sensitive learning based on CVaR and the robust-learning
interpretation of top-$k$ averaging.
\end{remark}

\begin{proposition}[Empirical RU minimizer and value]
Let $R_{(1)}(\theta)\le\cdots\le R_{(N)}(\theta)$ be the order statistics of $R_1,\dots,R_N$.
Fix $\alpha\in(0,1)$ and set $t:=(1-\alpha)N$, $m:=\lfloor t\rfloor$, $s:=t-m\in[0,1)$.
Consider
\begin{equation}
\widehat{\phi}_N(\eta) = \eta  +  \frac{1}{t}\sum_{i=1}^N (R_i-\eta)_+, 
\qquad (u)_+ := \max\{u,0\}.
\end{equation}
Then
\begin{equation}
\arg\min_{\eta\in\mathbb{R}} \widehat{\phi}_N(\eta)
=
\begin{cases}
\{R_{(N-m)}\}, & \text{if } s>0 \text{ and there are no ties at } R_{(N-m)},\\[2pt]
[R_{(N-m)},R_{(N-m+1)}], & \text{if } s=0 \text{ or there is a tie at the boundary.}
\end{cases}
\end{equation}
\end{proposition}

\begin{proof}
\textbf{1) Subgradient of $\widehat{\phi}_N$.}
For fixed $R_i$, the function $\eta\mapsto (R_i-\eta)_+$ is convex and piecewise linear with
\begin{equation}
\frac{d}{d\eta}(R_i-\eta)_+
=
\begin{cases}
-1, & \eta<R_i,\\
[-1,0], & \eta=R_i,\\
0, & \eta>R_i.
\end{cases}
\end{equation}
Summing and dividing by $t$,
\begin{equation}
\partial \widehat{\phi}_N(\eta)
=
\Big[1 - \tfrac{1}{t}\#\{i:R_i\ge \eta\}, 1 - \tfrac{1}{t}\#\{i:R_i> \eta\}\Big],
\label{eq:subgrad-interval}
\end{equation}
Where the sign $\#G$ means cardinality (number of element) in the  set G.
By convex optimality, $\eta^\star$ minimizes $\widehat{\phi}_N$ iff $0\in \partial \widehat{\phi}_N(\eta^\star)$, i.e.
\begin{equation}
\#\{i:R_i>\eta^\star\}\le t \le \#\{i:R_i\ge \eta^\star\}.
\label{eq:opt-count}
\end{equation}

\medskip
\textbf{2) What happens between consecutive order statistics.}
Fix an index $j\in\{0,1,\dots,N\}$ and consider $\eta\in\big(R_{(j)},R_{(j+1)}\big)$
(with the conventions $R_{(0)}=-\infty$, $R_{(N+1)}=+\infty$).
In this open interval, the active set is constant:
\begin{equation}
\#\{i:R_i>\eta\} = N-j, \qquad \#\{i:R_i\ge \eta\}=N-j.
\end{equation}
Hence the one-sided derivatives coincide and
\begin{equation}
\frac{d}{d\eta}\widehat{\phi}_N(\eta) = 1 - \frac{N-j}{t}
\qquad\text{for } \eta\in\big(R_{(j)},R_{(j+1)}\big).
\label{eq:local-slope}
\end{equation}
In particular, on the special interval $\big(R_{(N-m)},R_{(N-m+1)}\big)$ the slope equals
\begin{equation}
1 - \frac{m}{t} = \frac{s}{t}.
\label{eq:slope-special}
\end{equation}

\medskip
\textbf{3) Left and right behaviour at the boundary $R_{(N-m)}$.}
We will examine $\widehat{\phi}_N$ immediately to the \emph{left} and to the \emph{right} of $R_{(N-m)}$.

\emph{Right of $R_{(N-m)}$.}
By \eqref{eq:local-slope}-\eqref{eq:slope-special}, for every $\eta\in\big(R_{(N-m)},R_{(N-m+1)}\big)$
\begin{equation}
\frac{d}{d\eta}\widehat{\phi}_N(\eta) = \frac{s}{t}\ \ge\ 0.
\label{eq:right-slope}
\end{equation}
Thus $\widehat{\phi}_N$ is nondecreasing to the right of $R_{(N-m)}$ and strictly increasing there if $s>0$.

\emph{Left of $R_{(N-m)}$ (no ties case).}
If there are no ties at $R_{(N-m)}$, then for $\eta\in\big(R_{(N-m-1)},R_{(N-m)}\big)$ we have
$\#\{i:R_i>\eta\}=m+1$, so by \eqref{eq:local-slope}
\begin{equation}
\frac{d}{d\eta}\widehat{\phi}_N(\eta) = 1 - \frac{m+1}{t}
= \frac{s-1}{t} <0,
\label{eq:left-slope}
\end{equation}
because $s\in[0,1)$.

\emph{At the boundary with ties.}
Let $\vartheta:=R_{(N-m)}$ and denote $A:=\#\{i:R_i>\vartheta\}$ and $B:=\#\{i:R_i=\vartheta\}$.
By ranking, $A\le m$ (at most $m$ elements exceed $\vartheta$) and $B\ge 1$.
Since $t=m+s\in[m,m+1)$, we have
\begin{equation}
A \le m \le t \le m+1 \le A+B.
\end{equation}
Therefore the optimality condition \eqref{eq:opt-count} holds at $\eta=\vartheta$ even when $B>1$ (ties),
i.e. $0\in \partial \widehat{\phi}_N(\vartheta)$.

\medskip
\textbf{4) Minimizer set by cases.}

\emph{Case $s>0$ and no ties at $R_{(N-m)}$.}
By \eqref{eq:left-slope} $\widehat{\phi}_N$ is strictly decreasing just to the left of $R_{(N-m)}$,
and by \eqref{eq:right-slope} it is strictly increasing just to the right.
By convexity, the unique minimizer is the junction point:
\begin{equation}
\eta^\star = R_{(N-m)}.
\end{equation}

\emph{Case $s=0$ (integer tail size).}
Then \eqref{eq:right-slope} says the slope on $\big(R_{(N-m)},R_{(N-m+1)}\big)$ is $0$,
so $\widehat{\phi}_N$ is constant on that open interval.
Convexity and the left derivative $<0$ (as in \eqref{eq:left-slope} with $s=0$) imply that every point in the \emph{closed} interval
\begin{equation}
\eta^\star \in [R_{(N-m)},R_{(N-m+1)}]
\end{equation}
is a minimizer (a flat bottom).

\emph{Case ties at the boundary.}
From the boundary check above, $0\in\partial \widehat{\phi}_N(R_{(N-m)})$, so $R_{(N-m)}$ is a minimizer.
If, in addition, $R_{(N-m)}=R_{(N-m+1)}$ numerically, the set
$[R_{(N-m)},R_{(N-m+1)}]$ is a singleton equal to that tied value, hence also a valid description
of the minimizers in the statement.
(When $s=0$ and there are ties, the whole flat interval of minimizers remains valid.)

This completes the case analysis and the characterization of $\arg\min \widehat{\phi}_N$.
\end{proof}

\begin{proposition}[Empirical CVaR equals a fractional tail average (Proof of the proposition ~\eqref{prop_emp-cvar-fractional})]
Let $R_1,\dots,R_N\in\mathbb{R}$ and let $R_{(1)}\le \cdots \le R_{(N)}$ be the order statistics (ascending).
Fix $\alpha\in(0,1)$ and set $t:=(1-\alpha)N$, $m:=\lfloor t\rfloor$, $s:=t-m\in[0,1)$.
Consider the empirical Rockafellar-Uryasev objective
\[
\widehat{\phi}_N(\eta)=\eta+\frac{1}{(1-\alpha)N}\sum_{i=1}^N (R_i-\eta)_+ = \eta + \frac{1}{t}\sum_{i=1}^N (R_i-\eta)_+.
\]
Then its minimum value (the empirical CVaR) is
\begin{equation}
\label{eq:emp-cvar-fractional_}
\widehat{\mathrm{CVaR}}_\alpha(R)
:=\inf_{\eta\in\mathbb{R}}\widehat{\phi}_N(\eta)
=\frac{1}{t}\left(\sum_{i=N-m+1}^{N} R_{(i)} + sR_{(N-m)}\right).
\end{equation}
In particular, if $s=0$ (i.e., $t$ is an integer), then
\[
\widehat{\mathrm{CVaR}}_\alpha(R)=\frac{1}{t}\sum_{i=N-t+1}^{N} R_{(i)},
\]
the average of the top $t$ samples. If there is a tie at the boundary ($R_{(N-m)}=R_{(N-m+1)}$), the same Top-$k$ form holds with $k=\lceil t\rceil$.
\end{proposition}
\begin{proof}[Proof via the primal RU objective]
Let $t=(1-\alpha)N=m+s$ with $m=\lfloor t\rfloor$ and $s\in[0,1)$. 
Fix $\eta\in\big(R_{(N-m)},R_{(N-m+1)}\big)$, so exactly the top $m$ samples exceed $\eta$.
Then
\begin{equation}
\sum_{i=1}^N (R_i-\eta)_+ = \sum_{i=N-m+1}^{N} (R_{(i)}-\eta)
= \sum_{i=N-m+1}^{N} R_{(i)} - m\eta.
\end{equation}
Substituting in $\widehat{\phi}_N(\eta)=\eta+\frac{1}{t}\sum (R_i-\eta)_+$ gives the affine form
\begin{equation}
\widehat{\phi}_N(\eta) = \frac{1}{t}\sum_{i=N-m+1}^{N} R_{(i)} + \frac{s}{t}\eta,
\label{affine_form}
\end{equation}
so on this interval the slope is $s/t$.

\emph{Case 1: $s>0$ and no ties at $R_{(N-m)}$.}
For $\eta$ just below $R_{(N-m)}$, at least $m{+}1$ samples exceed $\eta$, so
\begin{equation}
\widehat{\phi}_N'(\eta+) = 1 - \frac{1}{t}\#\{i: R_i>\eta\} < 0.
\end{equation}
Immediately to the right, using the affine form ~\eqref{affine_form} of $\widehat{\phi}_N(\eta)$ we see that the slope is $s/t>0$. Hence $\widehat{\phi}_N$ decreases up to $\eta=R_{(N-m)}$ and increases after, so the unique minimizer is
\begin{equation}
\eta^\star = R_{(N-m)}.
\end{equation}

\emph{Case 2: $s=0$ (integer tail).}
Then the slope on $\big(R_{(N-m)},R_{(N-m+1)}\big)$ is $0$, hence $\widehat{\phi}_N$ is constant there. By convexity, every point in the closed interval
\begin{equation}
\eta^\star \in [R_{(N-m)},R_{(N-m+1)}]
\end{equation}
is a minimizer.

\emph{Case 3: Ties at the boundary.}
If $R_{(N-m)}=R_{(N-m+1)}$, the numeric interval degenerates, but the value attained at that boundary is the same; the (dual) fractional mass $s$ may be distributed across the tied block without changing the value.

In all cases, evaluating at $\eta=R_{(N-m)}$ yields
\begin{equation}
\inf_{\eta}\widehat{\phi}_N(\eta)
= \frac{1}{t}\sum_{i=N-m+1}^{N} R_{(i)} + \frac{s}{t}R_{(N-m)},
\end{equation}
which is the fractional tail average.
\end{proof}



\subsection{CVaR-consistency of the mean-squares RRA Formulation.}
\label{cvar_consistency_msq}
\begin{proposition}
Fix $\alpha\in(0,1)$ and  Let $(R_i)_{i=1}^N$ be nonnegative residual magnitudes, let $t:=\lfloor (1-\alpha)N \rfloor$ be an integer.
Consider $\varepsilon\ge 0$,  and the weights  $w_i=\tfrac{N}{t}\ge 0$ for $i\in\mathcal{I}_{\text{tail-}t}$ and 0 otherwise,  satisfying the following 
\begin{equation}
\frac{1}{N}\sum_{i=1}^N w_i = 1
\quad\text{(equivalently, }\sum_i w_i = N\text{).}
\label{eq:sumwN}
\end{equation}
Define the mean-squares robust penalty as
\(
\mathcal{P}_{\mathrm{ms}}(\theta,\varepsilon,w)
:=
\frac{1}{N}\sum_{i=1}^N w_i\big(R_i-\varepsilon\big)_{+}^{2}.\)
Then
\begin{equation}
\mathcal{P}_{\mathrm{ms}}(\theta,\varepsilon,w)
\ge
\big(\widehat{\mathrm{CVaR}}_{\alpha}(R)-\varepsilon\big)_+^{2}.
\end{equation}

\end{proposition}

\noindent\emph{Proof.}
Let $z_i := (R_i-\varepsilon)_+\ge 0$ and $\tilde w_i := w_i/N$ so that
$\tilde w_i\ge 0$ and $\sum_i \tilde w_i=1$ by \eqref{eq:sumwN}.
By Jensen's inequality for the convex map $x\mapsto x^2$,
\begin{equation}
\frac{1}{N}\sum_{i=1}^N w_iz_i^2
=
\sum_{i=1}^N \tilde w_iz_i^2
\ge
\Big(\sum_{i=1}^N \tilde w_iz_i\Big)^2
=
\Big(\frac{1}{N}\sum_{i=1}^N w_iz_i\Big)^2,
\end{equation}
Therefore, we have 
\begin{equation}
\mathcal{P}_{\mathrm{ms}}(R;\varepsilon,w)
\ge
\Bigg(\frac{1}{N}\sum_{i=1}^N w_i\big(R_i-\varepsilon\big)_{+}\Bigg)^{\!2}
\label{eq:ms-lower-bounds_1}
\end{equation}
Since $x\mapsto (x-\varepsilon)_+$ is convex and nondecreasing,
Jensen's inequality \citep{ruel1999jensen} yields
\begin{equation}
\frac{1}{N}\sum_{i=1}^N w_i(R_i-\varepsilon)_+
\ge
\Big(\frac{1}{N}\sum_{i=1}^N w_iR_i - \varepsilon\Big)_+,
\end{equation}
Therefore, we have 
\begin{equation}
\Bigg(\frac{1}{N}\sum_{i=1}^N w_i\big(R_i-\varepsilon\big)_{+}\Bigg)^{\!2}
\ge
\Bigg(\frac{1}{N}\sum_{i=1}^N w_iR_i-\varepsilon \Bigg) _{+}^{2}.
\label{eq:ms-lower-bounds_2}
\end{equation}
from  Eq. ~\eqref{eq:ms-lower-bounds_1} and ~\eqref{eq:ms-lower-bounds_2} we conclude that:
\begin{equation}
\mathcal{P}_{\mathrm{ms}}(\theta,\varepsilon,w)
\ge
\Bigg(\frac{1}{N}\sum_{i=1}^N w_iR_i-\varepsilon \Bigg) _{+}^{2},
\label{eq:partial_msq formulation}
\end{equation}

Fix $\alpha\in(0,1)$ and let $t:=\lfloor (1-\alpha)N \rfloor$ .
As defined in eq.~\eqref{cvar_descirption}, if  $w_i=\tfrac{N}{t}$ for $i\in\mathcal{I}_{\text{tail-}t}$ and $w_i=0$ otherwise, we have 
\begin{equation}
\frac{1}{N}\sum_{i=1}^N w_iR_i
=
\frac{1}{t}\sum_{i\in\mathrm{I}_{\text{tail-}t}} R_i
=
\widehat{\mathrm{CVaR}}_{\alpha}(R),
\end{equation}
Therefore,  \eqref{eq:partial_msq formulation} implies the bound
\begin{equation}
\mathcal{P}_{\mathrm{ms}}(\theta,\varepsilon,w)
\ge
\big(\widehat{\mathrm{CVaR}}_{\alpha}(R)-\varepsilon\big)_+^{2}.
\end{equation}
\hfill$\square$

\subsection{Statistical Interpretation of the  Empirical CVaR}
\label{subsec:stat-guarantees}
\paragraph{Interpretation of averaging the tail.}

Recall $R:=|r(X;\theta)|$ with $X\sim\mu$, and let $\mathrm{CVaR}_\alpha(R)$ be defined by the Rockafellar-Uryasev (RU) program
\begin{equation}
\mathrm{CVaR}_\alpha(R)
= \inf_{\eta\in\mathbb{R}} \left\{ \phi(\eta) := \eta + \frac{1}{1-\alpha}\mathbb{E}\big[(R-\eta)_+\big]\right\}.
\label{eq:RU-pop}
\end{equation}

At any minimizer $\eta^\star\in[q^-_\alpha,q^+_\alpha]$,
\[
\mathrm{CVaR}_\alpha(R)=\eta^\star + \frac{1}{1-\alpha}\mathbb{E}\big[(R-\eta^\star)_+\big],
\]
which is the average level of $R$ over the worst $(1-\alpha)$ fraction of outcomes. When the CDF is continuous at the quantile, this reduces exactly to the conditional mean $\mathbb{E}[R\mid R\ge \mathrm{VaR}_\alpha(R)]$; otherwise, it averages the strict upper tail and includes just enough of the atom at $\eta^\star$ to fill the $(1-\alpha)$ mass.
\subsection{Gradients and Subgradients}
\label{subsec:grads}

\begin{lemma}[Local constancy and measure-zero switches]
\label{lem:loc-const}
If $R_i(\theta)\neq R_j(\theta)$ for all $i\neq j$, then there exists a neighborhood $U$ of $\theta$ in which the permutation $\pi$ is constant and hence $\mathcal{I}_{\text{top-}k}(\theta)$ is constant. The set of $\theta$ where some $R_i(\theta)=R_j(\theta)$ or $R_i(\theta)=\varepsilon$ (or $r(x_i;\theta)=0$) is a finite union of co-dimension $\ge 1$ manifolds and thus has Lebesgue measure zero.
\end{lemma}

\begin{proof}[Idea]
Distinct order statistics remain distinct under small perturbations, so the sorting permutation is locally constant; equalities $R_i=R_j$, $R_i=\varepsilon$, $r=0$ define level sets of smooth maps.
\end{proof}

\subsubsection{CVaR violation penalty.}
Ignoring measure-zero tie sets (Lemma~\ref{lem:loc-const}), 
$\mathcal{I}_{\text{top-}k}$ is locally constant and for 
$\mathcal{P}_{\text{hinge}}(\theta,\varepsilon)
= \lambda_p \left( \widehat{\mathrm{CVaR}}_\alpha(R)- \varepsilon\right)_+^2$, we have 

\begin{equation}
\nabla_\theta \mathcal{P}_{\text{hinge}}(\theta,\varepsilon)
=2\cdot \mathbb{I}\{\widehat{\mathrm{CVaR}}_\alpha(R)>\varepsilon\} \big(\widehat{\mathrm{CVaR}}_\alpha(R)-\varepsilon\big) \cdot  \nabla_\theta R_i(\theta),
\quad
\nabla_\theta R_i(\theta) 
= \nabla_\theta |r(x_i;\theta)|.
\label{eq:cvar-grad}
\end{equation}

Whenever $r(x_i;\theta)\neq 0$,
\begin{equation}
\nabla_\theta |r(x_i;\theta)|
= \mathrm{sign}\!\big(r(x_i;\theta)\big)\nabla_\theta r(x_i;\theta).
\label{eq:abs-grad}
\end{equation}
At $r(x_i;\theta)=0$, the Clarke subdifferential is
\[
\partial_\theta |r(x_i;\theta)| = \big\{ s\nabla_\theta r(x_i;\theta) : s\in[-1,1] \big\},
\]
\paragraph{Gradients w.r.t. $\varepsilon$ and its parameter $\xi$.}With $\varepsilon=\psi(\xi)=\log(1+e^\xi)$ (softplus to ensure positivity),
we have 
$\frac{\partial \varepsilon}{\partial \xi}=\psi'(\xi)=\sigma(\xi)=\frac{1}{1+e^{-\xi}}.$ We have the following derivatives
\begin{equation}
\frac{\partial \mathcal{P}_{\text{hinge}}(\theta,\varepsilon)}{\partial \varepsilon}
= -2\cdot\mathbb{I}\{\widehat{\mathrm{CVaR}}_\alpha(R)>\varepsilon\}\big(\widehat{\mathrm{CVaR}}_\alpha(R)-\varepsilon \big).
\end{equation}
\begin{equation}
\frac{\partial \mathcal{P}_{\text{hinge}}(\theta,\varepsilon)}{\partial \xi}=\frac{\partial \mathcal{P}_{\text{hinge}}(\theta,\varepsilon)}{\partial \varepsilon}\cdot \frac{\partial \varepsilon}{\partial \xi}
= -2\cdot \mathbb{I}\{\widehat{\mathrm{CVaR}}_\alpha(R)>\varepsilon\}\bigg(\widehat{\mathrm{CVaR}}_\alpha(R)-\varepsilon \bigg)\cdot \sigma(\xi).
\end{equation}

\subsubsection{Squared-violation penalty.}
For $v_i(\theta,\varepsilon):=(R_i(\theta)-\varepsilon)_+$ and any fixed Top-$k$ selection,
\begin{equation}
\nabla_\theta \left[\frac{1}{k}\sum_{i\in\mathcal{I}_{\text{top-}k}} v_i^2\right]
= \frac{2}{k}\sum_{i\in\mathcal{I}_{\text{top-}k}}\mathbb{I}\{R_i>\varepsilon\}(R_i-\varepsilon)\nabla_\theta R_i.
\label{eq:penalty-theta-grad}
\end{equation}
At $R_i=\varepsilon$, take any subgradient in the interval between one-sided limits or smooth $(\cdot)_+$ (e.g., softplus/Huber).

\paragraph{Gradients w.r.t. $\varepsilon$ and its parameter $\xi$.}

\begin{equation}
\frac{\partial}{\partial \varepsilon}\left[\frac{1}{k}\sum_{i\in\mathcal{I}_{\text{top-}k}} v_i^2\right]
= -\frac{2}{k}\sum_{i\in\mathcal{I}_{\text{top-}k}}\mathbb{I}\{R_i>\varepsilon\}(R_i-\varepsilon),
\end{equation}


Hence,
\begin{equation}
\frac{\partial}{\partial \xi}\left[\frac{1}{k}\sum_{i\in\mathcal{I}_{\text{top-}k}} v_i^2\right]
= -\frac{2}{k}\left(\sum_{i\in\mathcal{I}_{\text{top-}k}}\mathbb{I}\{R_i>\varepsilon\}(R_i-\varepsilon)\right)\sigma(\xi).
\label{eq:penalty-eps-grad}
\end{equation}
Adding the regularizer $\gamma_\varepsilon\varepsilon$ contributes
\(
\partial_\xi(\gamma_\varepsilon\varepsilon)=\gamma_\varepsilon\sigma(\xi).
\)
\emph{We can confirm via the above derivation results  that the update of $\varepsilon$ is very sensible and depends on factors such as $\gamma_\varepsilon$and the values of the tail excess or individual excess.}

\subsection{Training Curves}
These  curves were are L2 relative error recorded every 500 epochs using  validation data sets randomly generated.
\subsubsection{1D Viscous Burger Equation}
\begin{figure}[ht!]
    \centering
    \includegraphics[width=0.75\linewidth]{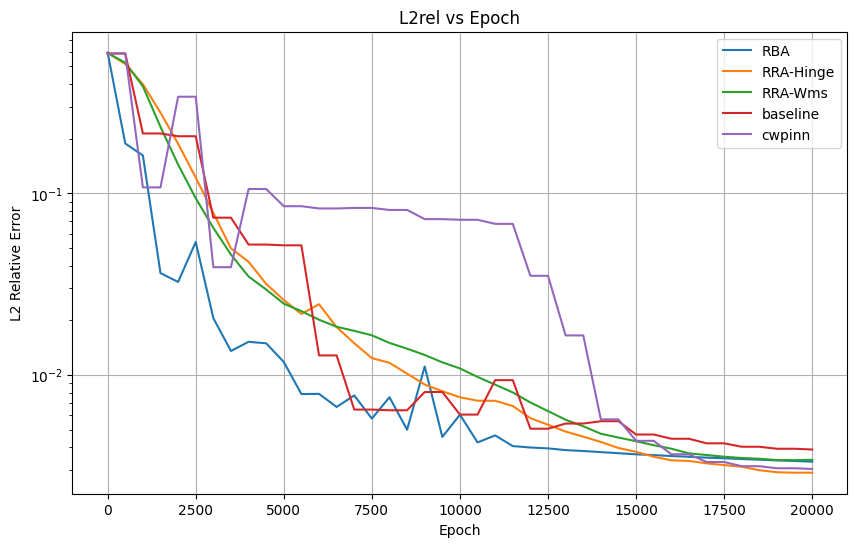}
    \caption{L2 Relative Error Training History for the 1D Viscous Burger Equation}
    \label{training_hist_burger}
\end{figure}

 The training history, including the evolution of the 
relative error, is presented in Figure~\eqref{training_hist_burger}. From this plot, we can clearly notice that the RRaPINN-based models have a more stable performance amelioration compare to  CWPINN, RBAPINN, and Baseline PINN.

\newpage
\subsubsection{1D Heat Equation}
The most stable learning curves appears to be that of RBA followed by the baseline. CWP reached a very low values very quickly but jumped unexpectedly later on and after get more stable towards the end. RRa-based models seems to have similar behavior with CWP, but achieve lower values toward the end when the robust tail manages to accomplish its duty.
\begin{figure}[ht!]
    \centering
    \includegraphics[width=0.75\linewidth]{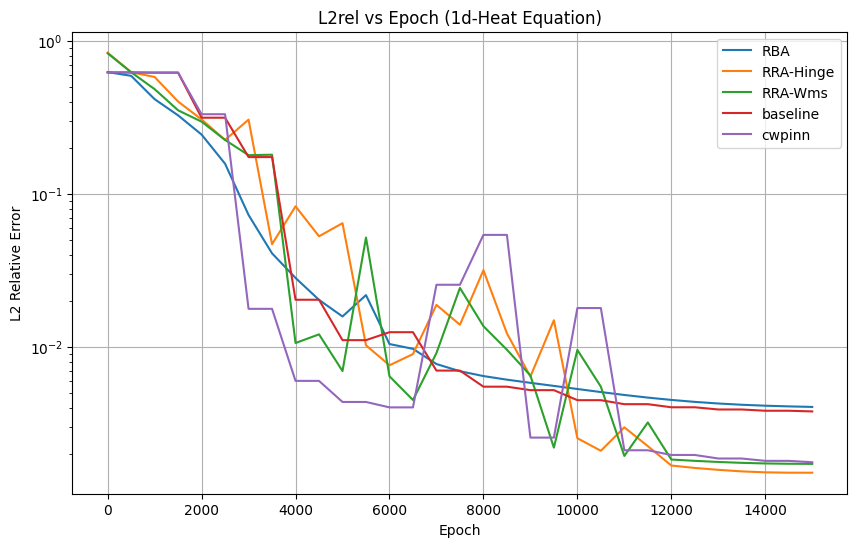}
    \caption{L2 Relative Error Training History for the 1D Heat Equation}
    \label{fig_1dheat}
\end{figure}

\subsubsection{KdV Equation}
All methods are very unstable before epoch 8k, and get stabilized after . RRaWMS and RBA have the most important decays, followed by RRa, CWP and Baseline PINNs.
\begin{figure}[ht!]
    \centering
    \includegraphics[width=0.75\linewidth]{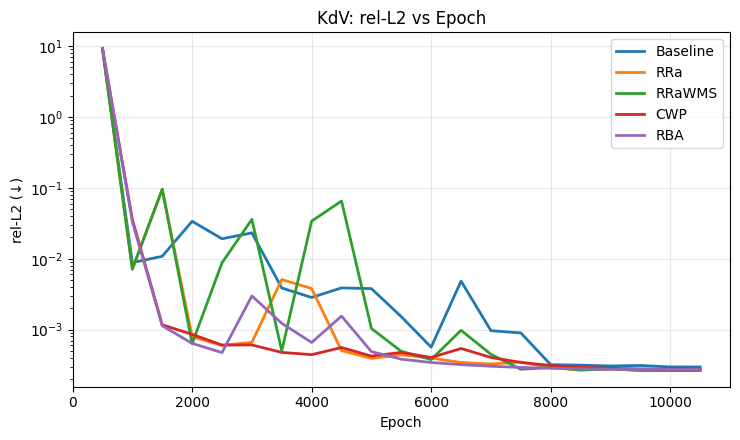}
    \caption{L2 Relative Error Training History  for the KdV Equation}
    \label{fig_kdv_eq}
\end{figure}

\subsubsection{2D Poisson Equation}
All methods are very unstable before epoch 11k, and get stabilized after . RRaWMS and RBA have the most important decays, followed by RRa, CWP and Baseline PINNs.
\begin{figure}[ht!]
    \centering
    \includegraphics[width=0.75\linewidth]{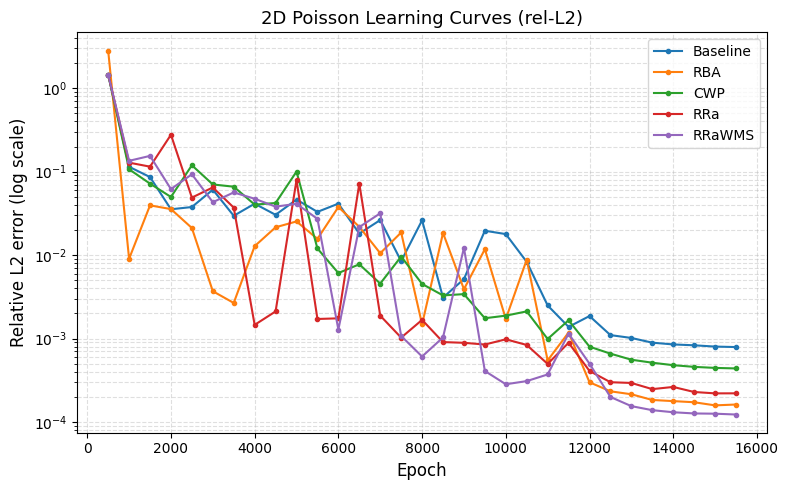}
    \caption{L2 Relative Error Training History  for the 2D Poisson Equation}
    \label{fig_2dpoisson}
\end{figure}
\newpage 
\subsection{Ablation study Curves}
\label{ablation_studies}
\subsubsection{CVaR Hinge penalty}
\begin{figure}[ht!]
  \centering
  \begin{subfigure}{0.48\linewidth}
    \includegraphics[width=\linewidth]{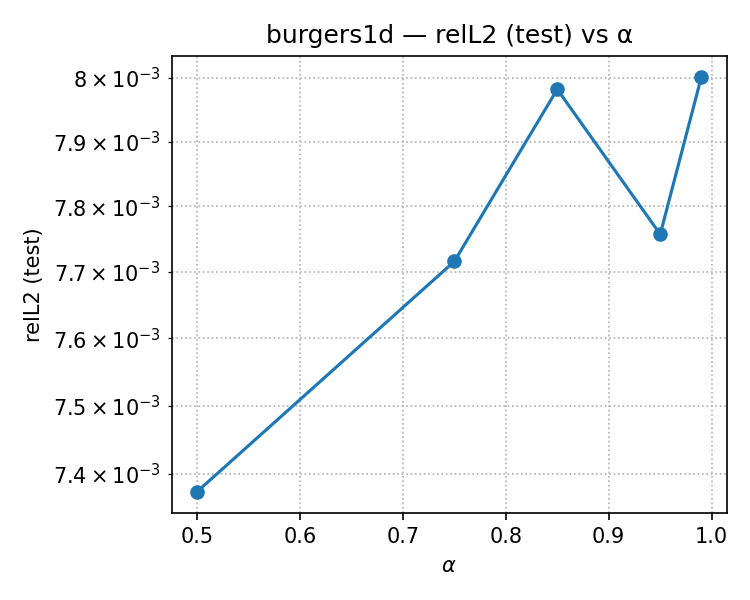}
    \caption{Relative $L_2$ vs.\ $\alpha$}
  \end{subfigure}
  \hfill
  \begin{subfigure}{0.48\linewidth}
    \includegraphics[width=\linewidth]{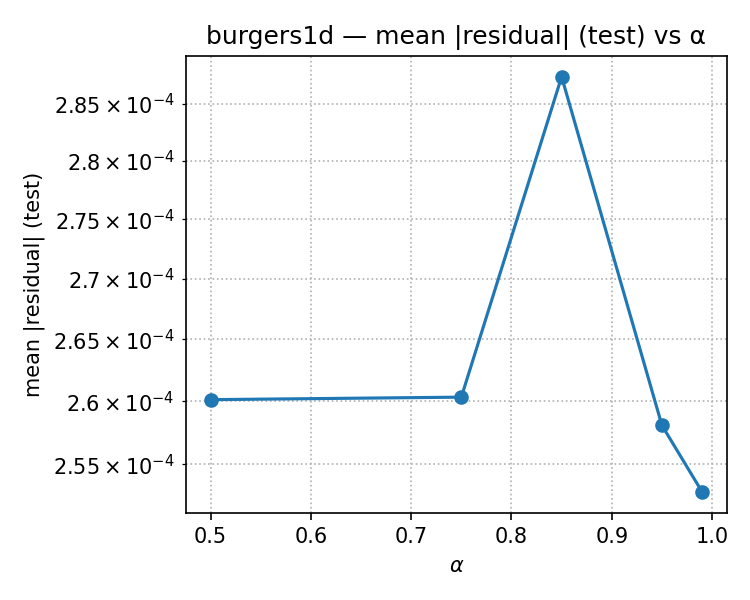}
    \caption{Mean absolute residual vs.\ $\alpha$}
  \end{subfigure}
  \caption{Ablation on $\alpha$ for the 1D viscous Burgers equation.}
  \label{fig_ablation-burgers_hinge}
\end{figure}

\begin{figure}[ht!]
  \centering
  \begin{subfigure}{0.48\linewidth}
    \includegraphics[width=\linewidth]{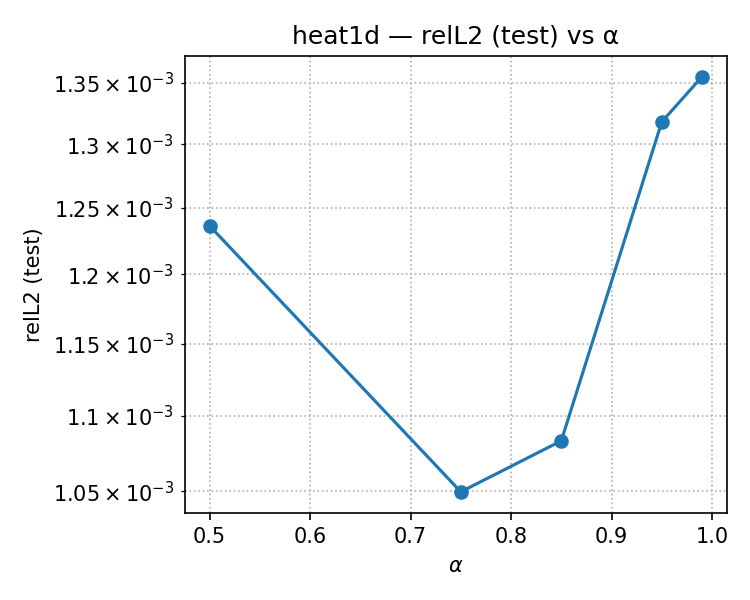}
    \caption{Relative $L_2$ vs.\ $\alpha$}
  \end{subfigure}
  \hfill
  \begin{subfigure}{0.48\linewidth}
    \includegraphics[width=\linewidth]{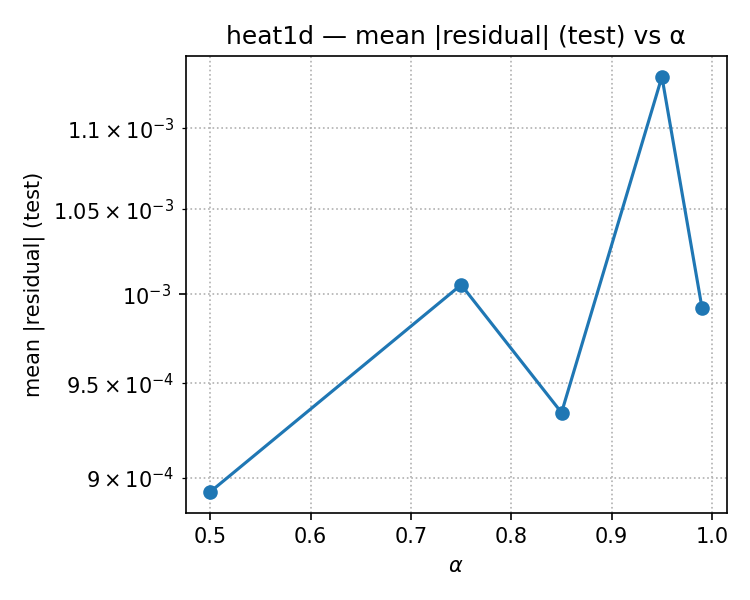}
    \caption{Mean absolute residual vs.\ $\alpha$}
  \end{subfigure}
  \caption{Ablation on $\alpha$ for the 1D heat equation.}
  \label{fig_ablation-heat_hinge}
\end{figure}

\begin{figure}[ht!]
  \centering
  \begin{subfigure}{0.48\linewidth}
    \includegraphics[width=\linewidth]{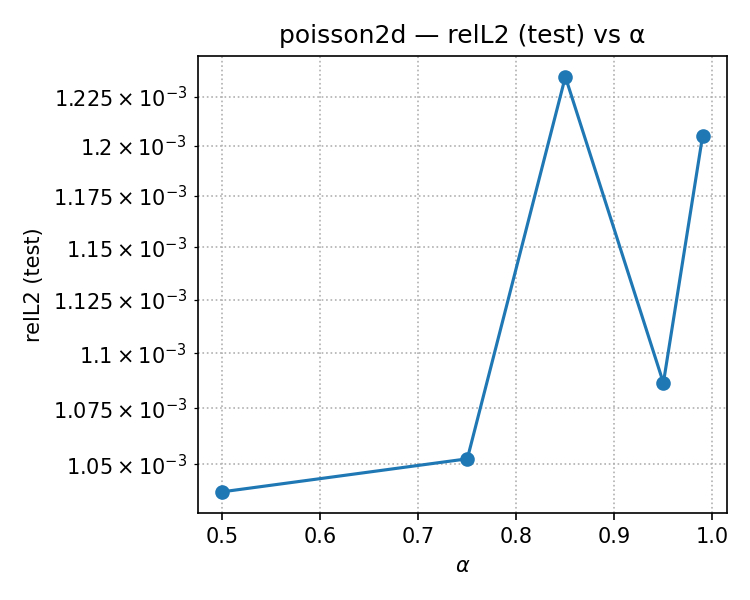}
    \caption{Relative $L_2$ vs.\ $\alpha$}
  \end{subfigure}
  \hfill
  \begin{subfigure}{0.48\linewidth}
    \includegraphics[width=\linewidth]{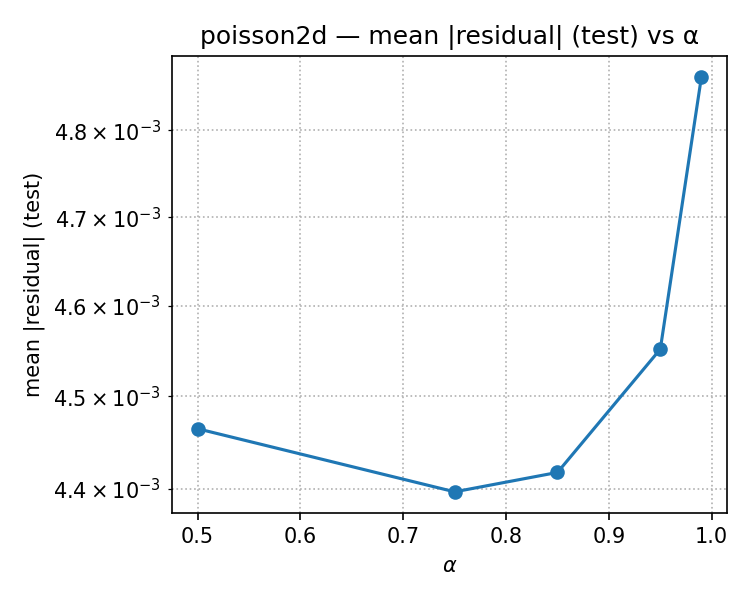}
    \caption{Mean absolute residual vs.\ $\alpha$}
  \end{subfigure}
  \caption{Ablation on $\alpha$ for the 2D Poisson equation.}
  \label{fig_ablation-poisson_hinge}
\end{figure}

\newpage
\subsubsection{Mean Excess penalty}

\begin{figure}[ht!]
  \centering
  \begin{subfigure}{0.48\linewidth}
    \includegraphics[width=\linewidth]{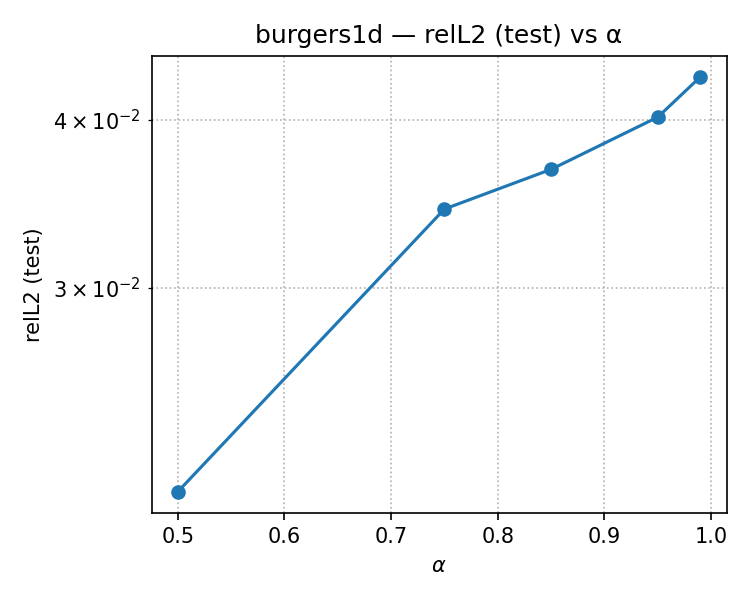}
    \caption{Relative $L_2$ vs.\ $\alpha$}
  \end{subfigure}
  \hfill
  \begin{subfigure}{0.48\linewidth}
    \includegraphics[width=\linewidth]{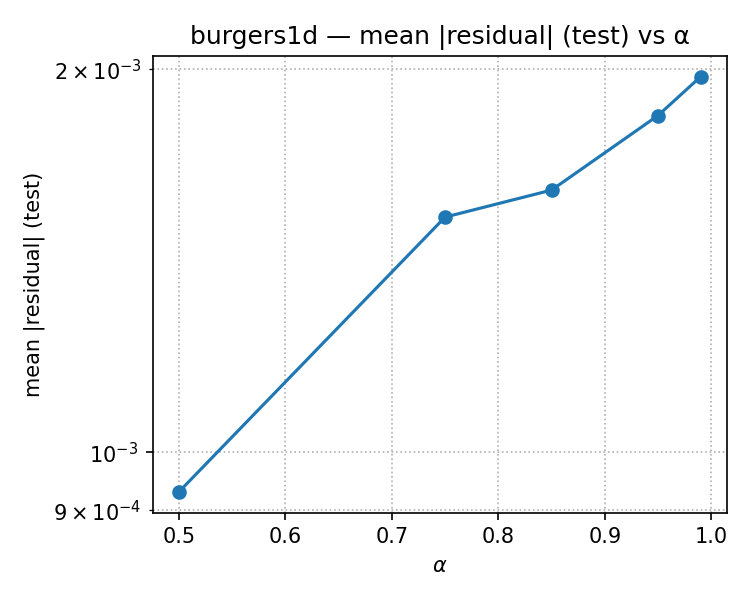}
    \caption{Mean absolute residual vs.\ $\alpha$}
  \end{subfigure}
  \caption{Ablation on $\alpha$ for the 1D viscous Burgers equation.}
  \label{fig_ablation-burgers}
\end{figure}

\begin{figure}[ht!]
  \centering
  \begin{subfigure}{0.48\linewidth}
    \includegraphics[width=\linewidth]{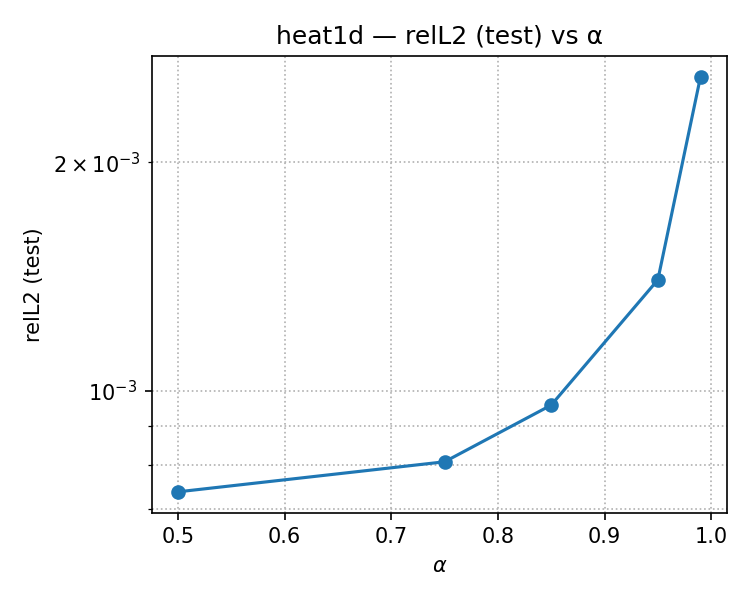}
    \caption{Relative $L_2$ vs.\ $\alpha$}
  \end{subfigure}
  \hfill
  \begin{subfigure}{0.48\linewidth}
    \includegraphics[width=\linewidth]{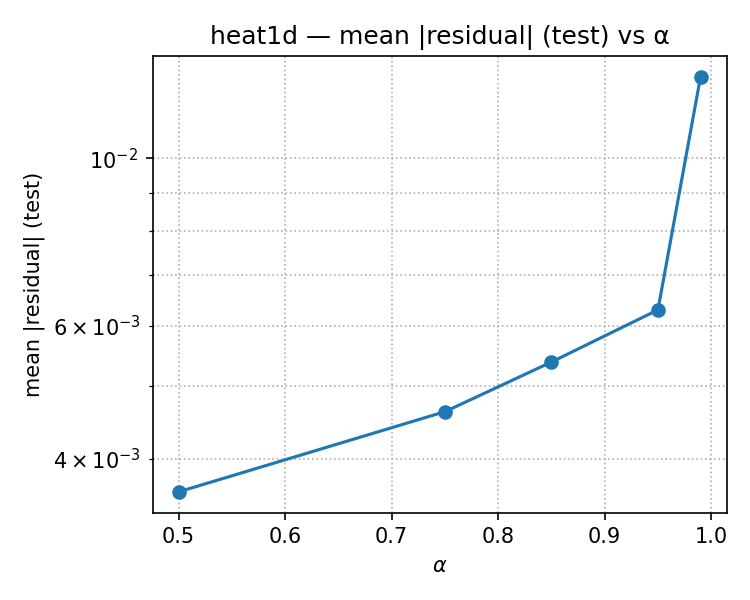}
    \caption{Mean absolute residual vs.\ $\alpha$}
  \end{subfigure}
  \caption{Ablation on $\alpha$ for the 1D heat equation.}
  \label{fig_ablation-heat}
\end{figure}

\begin{figure}[ht!]
  \centering
  \begin{subfigure}{0.48\linewidth}
    \includegraphics[width=\linewidth]{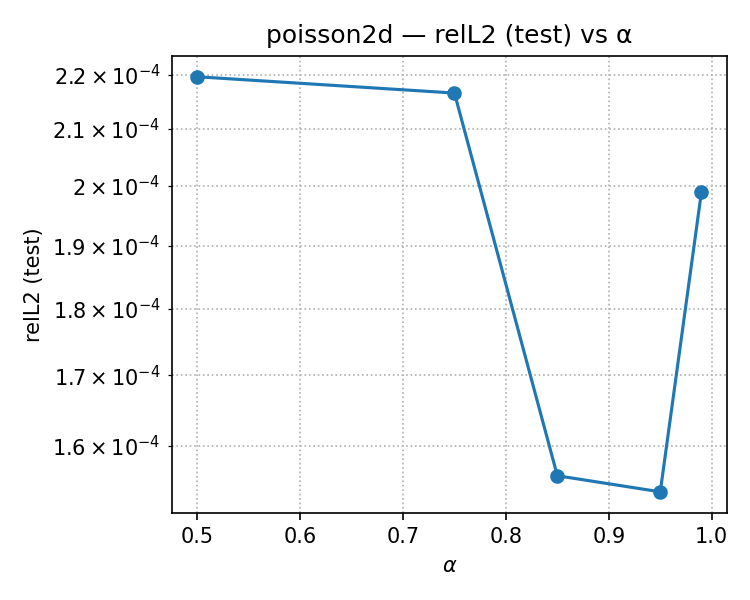}
    \caption{Relative $L_2$ vs.\ $\alpha$}
  \end{subfigure}
  \hfill
  \begin{subfigure}{0.48\linewidth}
    \includegraphics[width=\linewidth]{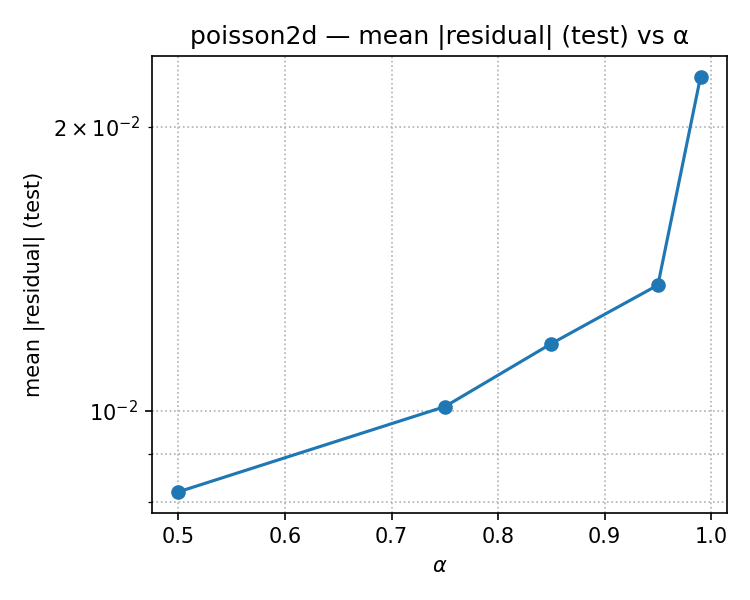}
    \caption{Mean absolute residual vs.\ $\alpha$}
  \end{subfigure}
  \caption{Ablation on $\alpha$ for the 2D Poisson equation.}
  \label{fig_ablation-poisson}
\end{figure}

\end{document}